\documentclass{article}

\usepackage{subcaption}
\usepackage{graphicx} 
\usepackage{xcolor}

\usepackage{amsmath}
\usepackage{amssymb}
\usepackage{mathtools}
\usepackage{xspace}

\usepackage{tikz}
\usepackage{pgfplots}
\usepgfplotslibrary{groupplots}
\usetikzlibrary{shapes.geometric, arrows, positioning, arrows.meta, calc, patterns, shapes.misc, matrix, backgrounds, external}
\pgfplotsset{compat=1.18}

\usepackage[round]{natbib}
\usepackage{cleveref}

\usepackage{enumerate}
\usepackage{enumitem}
\crefname{item}{}{}
\usepackage{url}

\newtheorem{corollary}{Corollary}

\newtheorem{lemma}{Lemma}
\newtheorem{proposition}{Proposition}

\let\parencite\citep
\let\textcite\citet

\title{Towards {U}nderstanding {E}poch-wise {D}ouble {D}escent in {T}wo-layer {L}inear {N}eural {N}etworks}
\author{Amanda Olmin, Fredrik Lindsten \\ Linköping University, Sweden}
\date{}

\usetikzlibrary{external}

\newcommand{\x}{x}
\newcommand{\y}{y}
\newcommand{\n}{n}
\newcommand{\xdim}{d_x}
\newcommand{\ydim}{d_y}

\newcommand{\Modelw}{W}

\newcommand{\Truew}{\overline{\Modelw}}
\newcommand{\modelz}{z}
\newcommand{\Modelz}{Z}
\newcommand{\truez}{\bar{\modelz}}
\newcommand{\Truez}{\bar{\Modelz}}
\newcommand{\starz}{\modelz^*}

\newcommand{\Layeri}[1]{\Modelw^{(#1)}}

\newcommand{\Layeriz}[1]{\Modelz^{(#1)}}
\newcommand{\yhat}{\hat{y}}
\newcommand{\h}{h}

\newcommand{\R}{\mathbb{R}}

\newcommand{\gaussian}{\mathcal{N}}

\newcommand{\xvecdd}{X} 
\newcommand{\yvecdd}{Y}
\newcommand{\epsvec}{E}
\newcommand{\V}{V} 
\newcommand{\U}{U} 
\newcommand{\Sx}{\Lambda}
\newcommand{\Syx}{\Sigma}
\newcommand{\lambdayx}{{\sigma}}
\newcommand{\lambdayxi}{{\sigma_i}}

\newcommand{\Vyx}{\V^{(yx)}}
\newcommand{\Uyx}{\U^{(yx)}}

\newcommand{\lr}{\eta}
\newcommand{\losstrain}{\mathcal{L}}

\newcommand{\lossgen}{\mathcal{L}_{\mathcal{G}}}
\newcommand{\lossgeni}{\mathcal{L}_{\mathcal{G}, i}}
\newcommand{\lossgenj}{\mathcal{L}_{\mathcal{G}, j}}

\newcommand{\activeset}{S_{\mathcal{A}}}

\newcommand{\rank}[1]{\text{rank}(#1)}
\newcommand{\trace}[1]{\text{Tr}\left(#1\right)}

\newcommand{\tht}{\text{th}}
\newcommand{\assump}{conditions (i)-(v)\xspace}

\newcommand{\myka}{\gamma^2 \lambda^2 + 4\lr^2 {\lambdayx}^2}
\newcommand{\mykai}{\gamma_i^2 \lambda_i^2 + 4\lr^2 {\lambdayxi}^2}
\newcommand{\mykb}{\gamma^2 + 4\lr^2 \modelz(t)^2 }
\newcommand{\diffz}{\big(\lambdayx - \lambda \modelz(t)\big)}
\newcommand{\diffzt}{\big(\truez - \modelz(t)\big)}
\newcommand{\mzt}{\modelz(t)}
\newcommand{\mzti}{\modelz_i(t)}

\newcommand{\mxt}{x(t)}

\newcommand{\az}{12 \lambda^2 \lr^2}
\newcommand{\bz}{-8\lr^2 \lambda \lambdayx (2-\rho)}
\newcommand{\cz}{2(\gamma^2 \lambda^2 + 2\lr^2 {\lambdayx}^2(1-\rho))}
\newcommand{\dz}{-\gamma^2 \lambda  \lambdayx(2-\rho)}

\newcommand{\ax}{12 \lambda^2 \lr^2}
\newcommand{\bx}{-4\lr^2 \lambda \lambdayx (7\rho-5)}
\newcommand{\cx}{2(\gamma^2 \lambda^2 + 2\lr^2 {\lambdayx}^2(5\rho^2-7\rho+ 2))}
\newcommand{\dx}{-\rho \lambdayx\Big(\gamma^2 \lambda + \frac{4\lr^2{\lambdayx}^2(1-\rho)^2}{\lambda}\Big)}

\newcommand{\eqspace}{\quad \qquad}

\newcommand{\expectation}{\mathbb{E}}

\begin{document}

\maketitle

\begin{abstract}
    Epoch-wise double descent is the phenomenon where generalisation performance improves beyond the point of overfitting, resulting in a generalisation curve exhibiting two descents under the course of learning. Understanding the mechanisms driving this behaviour is crucial not only for understanding the generalisation behaviour of machine learning models in general, but also for employing conventional selection methods, such as the use of early stopping to mitigate overfitting. While we ultimately want to draw conclusions of more complex models, such as deep neural networks, a majority of theoretical results regarding the underlying cause of epoch-wise double descent are based on simple models, such as standard linear regression. In this paper, to take a step towards more complex models in theoretical analysis, we study epoch-wise double descent in two-layer linear neural networks. First, we derive a gradient flow for the linear two-layer model, that bridges the learning dynamics of the standard linear regression model, and the linear two-layer diagonal network with quadratic weights. Second, we identify additional factors of epoch-wise double descent emerging with the extra model layer, by deriving necessary conditions for the generalisation error to follow a double descent pattern. While epoch-wise double descent in linear regression has been attributed to differences in input variance, in the two-layer model, also the singular values of the input-output covariance matrix play an important role. This opens up for further questions regarding unidentified factors of epoch-wise double descent for truly deep models. 

\end{abstract}

\section{Introduction}

The bias-variance trade-off implies a U-shaped generalisation error curve, as a function of model complexity. Accordingly, generalisation will improve with an increasing model complexity only to a certain point, after which the model will show signs of overfitting, and from which it never recovers. The double descent phenomenon contradicts this idea, demonstrating how generalisation performance can improve even beyond the point where the model perfectly fits the training data \parencite{Belkin2019, Nakkiran2021}, leading to a generalisation curve following a so-called double descent pattern. 

The double descent pattern in generalisation error has been observed not only with respect to model size, resulting in \textit{model-wise} double descent, but also in regards to the number of training epochs, when training models using iterative learning algorithms such as gradient descent. This type of double descent is known as \textit{epoch-wise} double descent, see e.g., \textcite{Baldi1991,Nakkiran2021}. While model-wise double descent has been studied extensively (\textcite{Belkin2019,Adlam2020,Advani2020,DAscoli2020,Hastie2022, Curth2023} to name a few), it typically involves understanding the asymptotic behaviour of models, assuming infinite training time. Meanwhile, epoch-wise double descent requires understanding the whole trajectory of learning. 

Commonly, epoch-wise double descent has been studied either mainly through empirical observation, e.g. \textcite{Nakkiran2021}, or through the theoretical analysis of simple models, such as linear regression \parencite{Heckel2021,Stephenson2021,Pezeshki2022} or the random feature model \parencite{Bodin2021, Bodin2022}. On the one end, \textcite{Nakkiran2021} connects epoch-wise double descent, similar to model-wise double descent, to the notion of \textit{model capacity}, attributing the second descent to the model capacity exceeding the number of training samples. On the other end, \textcite{Heckel2021,Pezeshki2022,Bodin2022} connect epoch-wise double descent to differences in speed of learning, describing the generalisation error as consisting of overlapping bias-variance trade-off curves, corresponding to fast- and slow-learned features. With this, the eigenvalues of the input covariance matrix are integral to the double descent phenomenon. Moreover, we do not need overparametrisation for the epoch-wise double descent pattern to emerge, something that is also consistent with earlier observations of the phenomenon, see e.g. \textcite{Baldi1991}. However, as the basis for these analyses are models consisting of just one layer of learnable parameters, while findings are corroborated empirically for deeper models, this leaves an open question of possibly unidentified factors causing epoch-wise double descent in deeper models. 

Although \textcite{Heckel2021} admittedly do also study double descent in a two-layer neural network with ReLU activation, the insights of \textcite{Heckel2021} do not connect double descent in two-layer networks directly to the properties of the data, but rather to the Jacobian of the network. Meanwhile, another line of work, studying the gradient dynamics of \textit{linear} two-layer neural networks, indicate that the singular values of the input-output covariance matrix of the training data are key factors in determining the speed of learning in these models \parencite{Saxe2014,Gidel2019,Lampinen2019,Tarmoun2021}. This suggests another possible contributor to the double descent phenomenon in two-layer linear neural networks, which has not been attributed as a factor of double descent in single-layer models. However, so far, and to the best of our knowledge, this has not been studied explicitly.

In this paper, we aim to start bridging the gap between epoch-wise double descent in the single-layer linear regression model and multi-layer neural networks, by studying double descent in the two-layer linear neural network. While this model is still linear, the training dynamics are, in contrast to the single-layer model, non-linear. Moreover, the benefits of studying linear models, is that we can obtain closed-form solutions for the training dynamics, at least in the case where weights are \textit{decoupled}, such that they evolve independently within a transformed space that is defined by the training data. More importantly, it opens up for the possibility of directly connecting epoch-wise double descent to properties of the training data. 

For our purposes, we derive the gradient flow for a two-layer diagonal, or decoupled, neural network, similar to what was studied by \textcite{Tarmoun2021}, but with non-isotropic input, and with individual learning rates for the first and second layer. With our formulation, we can recover the standard (single-layer) linear regression model as a special case of the dynamics, and hence we can connect our findings also to this model. For the two-layer decoupled linear neural network, we then
\begin{itemize}
    \item Derive the time-dependent generalisation error as a superposition of bias-variance curves, each corresponding to a single weight in the decoupled dynamics. The derivation is performed under the assumption that the true model is linear. 
    \item Characterise the behaviour of the individual bias-variance trade-off curves, and use this to find a necessary condition for epoch-wise double descent. 
\end{itemize}
Through this analysis, we identify additional factors of epoch-wise double descent in the two-layer model, not observed in the one-layer model, including a connection to the singular values of the input-output covariance matrix as well as an additional scenario under which epoch-wise double descent is possible.

In general, it is common to study linear shallow neural networks for gaining insights into the fundamental dynamics and behaviours of neural networks. Specifically related to the current work, are studies focusing on the learning dynamics of two-layer linear networks in gradient-based learning (apart from the ones mentioned above e.g., \textcite{Saxe2019,Pesme2021,Atanasov2022,Braun2022,Li2023,Berthier2023,Pesme2023,Even2023}), or the convergence of such learning algorithms (e.g., \textcite{Min2021,Xu2023}). Related are also studies on learning dynamics and convergence of multi-layer linear networks, see e.g., \textcite{Li2021b,Yun2021,Min2023,Chatterji2023}. 

In addition, the current work is closely connected to previous analyses of generalisation error in both linear and non-linear two-layer neural networks. This includes studies of asymptotic (in terms of the number of training epochs) generalisation error \parencite{Arora2019,Goldt2020}, and of the generalisation behaviour of two-layer models trained using \textit{one-pass} stochastic gradient descent, e.g.,  \textcite{Goldt2019,Mei2018,Goldt2020b}. Even more relevant are \textcite{Lampinen2019}, who derive the generalisation error for a two-layer linear network with isotropic input over the full course of learning. 
However, these previous works have not made direct connections to the epoch-wise double descent phenomenon. An exception is provided by \textcite{Wang2022}, who establish an upper bound on the generalisation error of two-layer neural networks based on a measure of \textit{gradient dispersion}. This measure, in turn, empirically demonstrates connections to epoch-wise double descent. Nevertheless, \textcite{Wang2022} do not cover an explicit theoretical analysis of the phenomenon.

\section{Preliminaries}
We will consider the problem of predicting the target $\y \in \R^{1 \times \ydim}$ given the input $\x \in \R^{1 \times \xdim}$, using a two-layer linear network with $\h$ hidden units
\begin{align*}
    &\yhat(\x) = \x \Modelw ^\top, \\
    & \Modelw = \Layeri{2} \Layeri{1},
\end{align*}
where $\Layeri{1} \in \R^{\h \times \xdim}$ and $\Layeri{2} \in \R^{\ydim \times \h}$. 

We train the linear network on $\n$ training data points $(x_i, y_i)$, $i=1, \ldots, \n$, using Mean Squared Error (MSE). Let 
\begin{align*}
    &\xvecdd = \begin{bmatrix}
        - \, \x_1 \, - \\
        - \, \x_2 \, - \\
        \cdots \\
        - \, \x_n \, - 
    \end{bmatrix}\in \R^{\n \times \xdim}, \quad \yvecdd = \begin{bmatrix}
        - \, \y_1 \, - \\
        - \, \y_2 \, - \\
        \cdots \\
        - \, \y_n \, - 
    \end{bmatrix} \in \R^{\n \times \ydim}, 
\end{align*}
be the input and output data matrices. The MSE criterion is
\begin{align}
    \losstrain = \frac{1}{2n} \trace { (\yvecdd - \xvecdd \Modelw^\top)^\top (\yvecdd - \xvecdd \Modelw^\top)}.
    \label{eq:full_loss}
\end{align}

Assuming small learning rates for each model layer, $\lr_a, \lr_b \geq 0$, we study the gradient flow
\begin{align}
    &   \frac{1}{\lr_a}\frac{d}{dt} \Layeri{1} = \frac{1}{n}{\Layeri{2}}^\top(\yvecdd^\top \xvecdd -  \Modelw \xvecdd^\top \xvecdd), \\
    &    \frac{1}{\lr_b}\frac{d}{dt} \Layeri{2} = \frac{1}{n} (\yvecdd^\top \xvecdd -  \Modelw \xvecdd^\top \xvecdd){\Layeri{1}}^\top,
    \label{eq:orig_dynamics}
\end{align}
resulting from training the two-layer linear network using gradient descent and MSE loss. 

\section{Theory}
In what follows, we derive decoupled dynamics for the two-layer linear neural networks, acting as a bridge between the dynamics of the one-layer linear model and the decoupled two-layer model with quadratic weights, studied by e.g., \textcite{Saxe2014,Gidel2019}. Then, we use these dynamics to study factors of epoch-wise double descent in two-layer linear networks.

\subsection{Decoupled dynamics of the two-layer linear network}

As a first step towards understanding double descent in two-layer linear neural networks, we will study the generalisation behaviour of the decoupled dynamics considered by e.g., \textcite{Saxe2014,Gidel2019,Lampinen2019,Tarmoun2021}. While a common assumption is $\xvecdd^\top \xvecdd = \mathbb{I}$, i.e the input is isotropic (see e.g., \textcite{Saxe2014, Lampinen2019, Tarmoun2021}), we instead assume a non-isotropic input where the input data matrix, $\xvecdd$, has the singular value decomposition (SVD) 
\begin{align*}
    n^{-1/2}\xvecdd = U\Lambda^{1/2}V^\top.
\end{align*}
The corresponding eigendecomposition of the input covariance matrix follows according to $n^{-1}\xvecdd^\top \xvecdd = \V \Lambda \V^\top$. Relaxing the assumption of an isotropic input, allows us to evaluate the effects of the eigenvalues of this covariance matrix on the learning dynamics. Alongside this, we assume that the input-output covariance matrix $n^{-1}\yvecdd^\top \xvecdd$ share right singular vectors with the input covariance matrix, and has the SVD
\begin{align*}
   & \n^{-1}  \yvecdd^\top \xvecdd = \Uyx \Syx {V^{(yx)}}^\top, \\
   & V^{(yx)} = V.
\end{align*}
Note that this assumption is similar to the one made by \textcite{Gidel2019}, with the exception that \textcite{Gidel2019} allow for the matrix $\Lambda$ to be non-diagonal, through the addition of a perturbation matrix, $B$. Here, we aim for simplicity and, hence, refer to \textcite{Gidel2019} for results on how the perturbation matrix $B$ might affect the final dynamics. Moreover, while our derivations should only require that the matrices $V$ and $V^{(yx)}$ share columns, up to a reshuffling of the elements of $\Lambda$ (see \cref{app:sec:decoupled_dynamics}), we keep it simple by assuming $V^{(yx)} = V$ in the main article. We note that, for the rank of the input-output covariance matrix $\yvecdd^\top \xvecdd$, we have $\rank{\yvecdd^\top \xvecdd}\leq \min\big(\rank{\yvecdd}, \rank{\xvecdd}\big)$, and therefore $\rank{\yvecdd^\top \xvecdd}\leq \rank{\xvecdd^\top \xvecdd}$.

Following e.g., \textcite{Saxe2014,Gidel2019,Lampinen2019,Tarmoun2021} we initialise weights using a \textit{spectral} initialisation according to $\Modelw(0) = \Uyx \Modelz(0) \V^\top$, where $\Modelz = \Layeriz{2} \Layeriz{1}$, and with $\Layeriz{1} \coloneq \Layeri{1}V$, $\Layeriz{2} \coloneq {\Uyx}^\top \Layeri{2}$. Then, we study the evolution of what \textcite{Saxe2014} termed the \textit{synaptic} weights $\Layeriz{1}, \Layeriz{2}$
\begin{align}
    &   \frac{1}{\lr_a}\frac{d}{dt} \Layeriz{1} = {\Layeriz{2}}^\top({\Syx} - \Modelz \Lambda), \nonumber \\
    &    \frac{1}{\lr_b}\frac{d}{dt} \Layeriz{2} = ({\Syx} - \Modelz \Lambda ) {\Layeriz{1}}^\top.
    \label{eq:synaptic_dynamics}
\end{align}
Observe that as $\Uyx$ and $\V$ are constant matrices, we will keep the relationship between $\Modelw$ and $\Modelz$ throughout learning. Moreover, we follow \textcite{Saxe2014,Gidel2019,Lampinen2019,Tarmoun2021} and \textit{decouple} weights by selecting $\Modelz(0)$ to be diagonal. This is done by initialising the $i^{\tht}$ column, $\alpha^{(i)}$, of $\Layeriz{1}$, and the $i^{\tht}$ row, ${\beta^{(i)}}^\top$, of $\Layeriz{2}$ according to $\alpha^{(i)}, \beta^{(i)} \, \propto \, r^{(i)}$, with $r^{(i)}$ a constant, unit vector. Importantly, $r^{(i)}$, for $i=1, \ldots, \min(\xdim, \ydim)$, are chosen to be orthogonal, i.e. $r^{(i)} \cdot r^{(j)} = 0$ for $i \neq j$. Decoupling weights in this manner, the rank of $\Modelz$ will be  upper-bounded by $\h$. In the case of an undercomplete hidden layer, i.e. $\h < \min (\xdim, \ydim)$, this requires that we initialise some diagonal weights to $0$. We initialise diagonal weight $i$ at $0$ by setting $r^{(i)}$ equal to the vector of zeroes, with $r^{(i)}  = \mathbf{0}$. 

With the given initialisation, $\Modelz$ will remain diagonal, see e.g. \textcite{Saxe2014} and \cref{app:sec:decoupled_dynamics}, resulting in the decoupled two layer dynamics where weights along the diagonal evolve independently. In the decoupled dynamics, a single weight along the diagonal of $\Modelz$ can be described by $\modelz_i= \alpha^{(i)} \cdot \beta^{(i)}=a_ib_i$, with scalar projections  $a_i = \alpha^{(i)} \cdot r^{(i)}, b_i = \beta^{(i)} \cdot r^{(i)}$ evolving as
\begin{align}
    \frac{1}{\lr_a}\frac{da_i}{dt} = b_i(\lambdayxi - \lambda_i a_ib_i), \nonumber \\
    \frac{1}{\lr_b}\frac{db_i}{dt} = a_i(\lambdayxi - \lambda_i a_ib_i).
    \label{eq:two_layer_dynamics_a_b_relaxed}
\end{align}
We let $\lambda_i$ and $\lambdayxi$ be the $i^{\tht}$ diagonal elements of $\Lambda$ and $\Syx$, respectively. Observe that, $\lambda_i, \lambdayxi \geq 0$ by definition.

To obtain a gradient flow for the product $\modelz_i=a_i b_i$, 
we again follow previous work, e.g. \textcite{Tarmoun2021}, and make use of conserved quantities of the dynamical system \cref{eq:two_layer_dynamics_a_b_relaxed}. Specifically, with the relaxed assumption of different learning rates, solutions to \cref{eq:two_layer_dynamics_a_b_relaxed} follow trajectories for which $\lr_b a_i^2 - \lr_a b_i^2  = \gamma_i$, where $\gamma_i$, for $i=1, \ldots, \min(\xdim,\ydim)$, is a constant (see \cref{app:sec:bridged_dynamics} for the proof). Then,
\begin{align}
    \frac{d\modelz_i}{dt} &= 
    \sqrt{\gamma_i^2 + 4 \lr^2 \modelz_i^2} (\lambdayxi - \lambda_i \modelz_i),
    \label{eq:two_layer_dynamics_z_relaxed}
\end{align}
with $\lr \coloneqq \sqrt{\lr_a \lr_b}$. This is an extension of the gradient flow studied by \textcite{Tarmoun2021} with what is referred to as an asymmetric, spectral initialisation. Here, we extend the dynamics of \textcite{Tarmoun2021} by allowing for different learning rates in the two layers as well as allowing $\lambda_i$ to be different from $1$. We note that, concurrent with this manuscript, \textcite{Kunin2024} published a similar gradient flow to ours, using different learning rates in the two model layers. However, the gradient flow of \textcite{Kunin2024} does not depend on a decoupling of the weights, and hence a solution to the dynamics is provided only in the special case $\h=1$ and $\lambda_i=1$. Here, we provide a solution to the decoupled dynamics for general $\h$ and $\lambda_i$, although we focus on an initialisation upper bounded according to $\modelz_i(0) < \lambdayxi / \lambda_i$, which is of relevance for our later analysis.
\newpage
\begin{proposition}
Consider $\modelz_i(t)$ initialised at $\modelz_i(0) < \lambdayxi / \lambda_i$ and assume $\lambda_i > 0$. If $\gamma_i \neq 0$, the solution to \cref{eq:two_layer_dynamics_z_relaxed} is
\begin{align}
    \modelz_i(t) = \frac{C^2 \lambda_i^2 \lambdayxi e^{2\sqrt{ \mykai } t}-2 C \gamma_i^2 \lambda_i^2 e^{\sqrt{ \mykai } t}-4 \gamma_i^2 \lr^2 \lambdayxi}{\lambda_i \left(C^2 \lambda_i^2 e^{2\sqrt{ \mykai } t} + 8 C  \lr^2 \lambdayxi e^{\sqrt{ \mykai } t}-4 \gamma_i^2\lr^2\right)},
    \label{eq:z_dyn_bridged_sol}
\end{align}
with
\begin{align*}
    C = \frac{\sqrt{\Big(\mykai\Big)\Big(\gamma_i^2 + 4 \lr^2 \modelz_i(0)^2 \Big)}+ \gamma_i^2 \lambda_i + 4  \lr^2 \lambdayxi \modelz_i(0)}{\lambda_i \big(\lambdayxi - \lambda_i \modelz_i(0)\big)}.
\end{align*}
Moreover,
\begin{align}
    \lim_{t\rightarrow \infty} \mzti = \frac{\lambdayxi}{\lambda_i},
    \label{eq:convergence_zi}
\end{align}
and the weight $\modelz_i(t)$ converges to the point $\starz_i = \lambdayxi / \lambda_i$ at a rate $\mathcal{O}\Big(e^{-\sqrt{ \mykai } t}\Big)$. If instead $\gamma_i = 0$, the above (\cref{eq:z_dyn_bridged_sol,eq:convergence_zi}) holds provided that $\lr > 0$ and $\modelz_i(0) > 0$. 
\label{prop:dyn_sol}
\end{proposition}

We refer to \cref{app:sec:dyn_sol}, for the proof of \cref{prop:dyn_sol}. We observe that the convergence rate is dependent on both the eigenvalue $\lambda_i$ and the singular value $\lambdayxi$, where the parameters $\gamma_i, \lr$ control their respective influence. For a large absolute value of the conservation quantity $\gamma_i$ (also referred to as the level of initialisation imbalance, see e.g. \textcite{Tarmoun2021}), relative to the parameter $\lr$, the eigenvalue $\lambda_i$ of the input covariance matrix will dominate the rate of convergence. If instead the learning rate parameter $\lr$ is large, relative to $|\gamma_i|$, the singular value $\lambdayxi$ of the input-output covariance matrix will dominate this rate.   
 
Throughout the paper, we will pay special attention to two special cases of the dynamics. The first special case is $\lr_a=0$ ($\lr=0$), which corresponds to fixed first layer weights, resulting in the dynamics of the standard linear regression model, see e.g. \textcite{Gidel2019}, with solution 
\begin{align}
    \modelz_i(t) = e^{-|\gamma_i| \lambda_i  t} \big(\modelz_i(0) - \lambdayxi\lambda_i^{-1}\big) + \lambdayxi\lambda_i^{-1}.
    \label{eq:one_layer_solution_z}
\end{align}
We will sometimes also refer to the linear regression model as the \textit{one-layer} model. Importantly, since we recover the one-layer model as a special case of the decoupled dynamics, we will be able to connect our findings to previous results on epoch-wise double descent for this model (e.g., \textcite{Heckel2021,Pezeshki2022}).

The second special case is $\gamma_i=0$, for which we recover the \textit{balanced} dynamics, studied by e.g., \textcite{Saxe2014, Gidel2019}, with solution
\begin{align}
    \modelz_i(t) = \frac{\lambdayxi e^{2\lr \lambdayxi t} z_i(0)}{ \lambda_i (e^{2\lr \lambdayxi t}  - 1)z_i(0) + \lambdayxi}.
    \label{eq:two_layer_solution_z}
\end{align}
We observe that the two special cases represents boundary cases of the general solution, in the sense that for $\lr=0$, the time-dependent exponential term depends only on the eigenvalue value $\lambda_i$, and not on $\lambdayxi$, while for $\gamma_i=0$, the time dependent exponential term instead depends only on $\lambdayxi$, and not on $\lambda_i$.

We note that under \cref{eq:two_layer_dynamics_z_relaxed}, weights $\modelz_i(t)$ that are initialised at $0$, by initialising $a_i(0)=b_i(0)=0$ (resulting in $\gamma_i=0$), will remain at $0$. Hence, depending on the initialisation, only a subset of the diagonal elements of $\Modelz(t)$ will change during the course of learning. We will refer to these non-constant weights as \textit{active} and use $\activeset \subseteq \{1, 2, \ldots, \min(\xdim, \ydim)\}$ with size $|\activeset|$ to denote the set of indices corresponding to active weights in $\Modelz(t)$.

\subsection{Epoch-wise double descent in the two-layer model}
We study the generalisation dynamics of $\Modelz$, when weights are learned following the decoupled dynamics in  \cref{eq:two_layer_dynamics_z_relaxed}. First, for deriving the test MSE, we will assume that $x$ is Gaussian with mean zero and true covariance matrix $\bar{\Lambda}$, written as $x \sim \gaussian(\mathbf{0}, \bar{\Lambda})$. In addition, we assume that $y$ follows a linear model
\begin{align}
    \y = \x \Truew^\top + \epsilon,
    \label{eq:true_model}
\end{align}
with true weight matrix $\Truew \in \R^{\ydim \times \xdim}$. The noise parameter, $\epsilon \in \R^{1 \times \ydim}$, is assumed to be zero-mean Gaussian with covariance matrix $\Lambda^{(\epsilon)}$, i.e. we assume $\epsilon \sim \gaussian(\mathbf{0}, \Lambda^{(\epsilon)})$. In addition, we assume this parameter to be independent of $\x$. 

In the synaptic weight space, the true weight matrix (as used in \cref{eq:true_model}) is 
\begin{align}
    \Truez \coloneqq {U^{(yx)}}^\top \Truew \V. 
    \label{eq:true_weights_synaptic}
\end{align}
We provide an alternative formulation of $\Truez$ in the following lemma. 

\begin{lemma}
    Assume that elements of the output data matrix, $\yvecdd$, follow the linear model given in \cref{eq:true_model} and let 
    \begin{align*}
        \epsvec = \begin{bmatrix}
            - \, \epsilon_1  \, - \\
            - \, \epsilon_2  \, - \\
            \cdots \\
            - \, \epsilon_n \, - 
        \end{bmatrix},
    \end{align*}
    be the matrix of residual vectors $\epsilon_i = \y_i - \x_i \Truew^\top$, for $i = 1, \ldots, n$. The true synaptic weight matrix, given in \cref{eq:true_weights_synaptic}, can be rewritten according to
    \begin{align*}
        &\Truez = \Syx \Lambda^{\dagger} - \widetilde{\epsvec}^\top ({\Lambda^{1/2}})^{\dagger}, \\
        &\widetilde{\epsvec} \coloneqq n^{-1/2}{U^{(yx)}}^\top \epsvec U,
    \end{align*}
    with $\dagger$ denoting the Moore-Penrose pseudoinverse.

    Subsequently, the $i^{th}$ diagonal element of $\Truez$, assuming that $\lambda_i > 0$, can be expressed as
    \begin{align}
        \truez_i = \lambdayxi \lambda_i^{-1} - \tilde{\epsilon}_i \lambda_i^{-1/2}.
        \label{eq:single_true_weight}
    \end{align}
    where we let $\tilde{\epsilon}_i$ denote the $i^{th}$ element on the main diagonal of $ \widetilde{\epsvec}$.
    \label{lemma:true_synaptic_weight_expr}
\end{lemma}

For the purpose of studying epoch-wise double descent, we use \cref{lemma:true_synaptic_weight_expr} to derive a time-dependent expression for the generalisation error of the decoupled two-layer linear network. 

\begin{proposition}
    Consider the diagonal weight matrix $\Modelz(t)$ with weights $\modelz_i(t)$ on the diagonal following the decoupled dynamics in \cref{eq:two_layer_dynamics_z_relaxed}. Moreover, assume $x \sim \gaussian(\mathbf{0}, \bar{\Lambda})$ and that $y$ follows the linear model given by \cref{eq:true_model}. Then, using MSE as the error metric and approximating $\bar{\Lambda} \approx \V \Lambda \V^\top$, the total generalisation error of $\Modelz(t)$ can be written as a sum of $|\activeset|$ individual error curves, according to
    \begin{align}
        \lossgen(t) &= \frac{1}{2}\expectation_{\x, \y} \bigg[ (\y - \x \Modelw(t)^\top) (\y - \x \Modelw(t)^\top)^\top \bigg] \nonumber \\ & \approx \frac{1}{2}\sum_{i \in \activeset} \lambda_i \big(\truez_i - z_i(t)\big)^2  + \text{const.}
        \label{eq:full_test_loss_z}
    \end{align}
    The true weights, $\truez_i$, are given by \cref{lemma:true_synaptic_weight_expr}. Each individual error term within the sum in \cref{eq:full_test_loss_z} is either monotonically decreasing, U-shaped, or monotonically increasing, and, hence, the total generalisation error is a sum of such curves. 
    \label{prop:test_mse}
\end{proposition}

Following \cref{prop:test_mse}, the total generalisation error of $\Modelz(t)$ is described by $|\activeset|$ overlapping error curves. Depending on the initialisation, the test MSE might include additional constant error, covering error corresponding to inactive diagonal elements, i.e., $\modelz_i(t)$ for which $i \notin \activeset$. Constant error can also arise from assuming that the weight matrix, $\Modelz(t)$, is diagonal, while the true weight $\Truez$ may not be. Moreover, we bring attention to the fact that \cref{eq:full_test_loss_z} relies on the assumption that the training sample size is large enough for the sample covariance matrix,  $V \Lambda \V^\top$, to approximate the true covariance matrix, $\bar{\Lambda}$ (i.e. $\bar{\Lambda} \approx \V \Lambda \V^\top$). Proofs for both \cref{lemma:true_synaptic_weight_expr,prop:test_mse} are provided in \cref{app:sec:test_mse}.

We acknowledge that similar expressions for the generalisation error, as given by \cref{prop:test_mse}, have been provided previously. In comparison with \textcite{Lampinen2019}, we consider the more general decoupled dynamics given by \cref{eq:two_layer_dynamics_z_relaxed}, while \textcite{Lampinen2019} consider the balanced dynamics as well as an isotropic input, i.e. $\lambda_i=1$ and $\gamma_i=0 \, \forall i \in [1, \ldots \min(\xdim, \ydim)]$. In this context, we also mention \textcite{Heckel2021}, who determine a time-dependent upper bound on the test MSE of a two-layer neural network with ReLU activation. To derive this upper bound, \textcite{Heckel2021} assume the neural network to be wide (i.e. the hidden dimension $\h$ to be large), and utilise a kernel approximation of the network. According to \textcite{Heckel2021}, the final expression can be broadly interpreted as a sum over $\n$ bias-variance trade-off curves.  Moreover, the general expression in \cref{eq:full_test_loss_z}, without making assumptions regarding the dynamics of the diagonal weights $\modelz_i(t)$, naturally also shares similarities with corresponding expressions derived for the one-layer and random feature models when training with (stochastic) gradient descent, see e.g., \textcite{Advani2020,Heckel2021,Pezeshki2022,Bordelon2022}.

Indeed, for the decoupled two-layer neural network, we uncover additional commonalities with the one-layer model. Namely, with each diagonal weight $\modelz_i(t)$ following the decoupled two-layer dynamics in \cref{eq:two_layer_dynamics_z_relaxed}, we find that a single curve in the sum of \cref{eq:full_test_loss_z} will not, by itself, exhibit a double descent pattern.  Therefore, and equivalent to what \textcite{Heckel2021} found for the one-layer model, epoch-wise double descent in the decoupled linear two-layer model, must be a result of overlapping error curves. Moreover, each such error curve corresponds to the error of a single weight, $\modelz_i(t)$. This contrasts with the findings of \textcite{Heckel2021} for the non-linear two-layer model, where the upper bound on the generalisation error is derived from a singular value decomposition of the Gram matrix of the kernel approximation, and the bias-variance curves correspond to dimensions of this Gram matrix. 

\subsubsection{Analysis of error curves}

To understand which factors give rise to the double descent behaviour, we start by analysing the individual error curves in \cref{eq:full_test_loss_z}. First, observe that the initialisation $\modelz_i(0)$ as well as the position of the true minimum $\truez_i$ relative to $\modelz_i(0)$ and the global minimum $\modelz_i^*$, will determine the behaviour of $\modelz_i(t)$, as well as the shape of its error curve. For example, if $\modelz_i(0) > \lambdayxi / \lambda_i$, $\modelz_i(t)$ will decrease in $t$, while for the opposite ($\modelz_i(0) < \lambdayxi / \lambda_i$), $\modelz_i(t)$ will increase in $t$. Moreover, if $\truez_i$ does not lie on the path between the initialisation $\modelz_i(0)$ and the global minimum $\modelz_i^*$, then $\modelz_i(t)$ will never pass this point and so, the corresponding error curve will be either monotonically decreasing or monotonically increasing in $t$. 

To limit the number of possible scenarios, we will restrict our analysis to the case where $\modelz_i(t)$ is positive and increasing in $t$. In addition, we will assume that $\truez \in [0, \modelz_i^*]$, noting that this also covers cases where the error curve of $\modelz_i(t)$ is monotonically increasing/decreasing in $t$, albeit with a focus on what we presume to be the most interesting scenario; the one where $\truez_i \in [\modelz_i(0), \modelz_i^*]$, such that the error curve for $\modelz_i(t)$ is U-shaped. Hence, for our further analysis, we will assume that the following two conditions hold for each $i \in \activeset$:
\begin{enumerate}[label=(\roman*)]
    \item The model weight $\modelz_i(t)$ is initialised with $\modelz_i(0) \in [0, \truez_i]$.
    \item The true minimum $\truez_i$ lies in the interval [0, $\modelz_i^*]$, such that it can be reparameterised as $\truez_i = (1-\rho_i) \modelz_i^*$, where $\rho_i$ is a constant in the interval $[0, 1]$. 
    \label{it:last_assumption}
\end{enumerate}
Condition (i) ensures that each active weight $\modelz_i(t)$ is increasing in $t$ and remains positive throughout training. Except for initialising $\modelz_i(t) \geq 0$, we put an upper bound $\truez_i \geq \modelz_i(0)$, such that the weight $\modelz_i(t)$ is initialised before or at the true minimum, $\truez_i$. Meanwhile, condition (ii) restricts $\truez_i$ to lie on the path between $0$ and $\modelz_i^*$. With this restriction, we put a focus on the scenarios where the error curve for $\modelz_i(t)$ is U-shaped, while still including scenarios where the error curve is monotonically decreasing in $t$ ($\rho_i = 0$) as well as monotonically increasing in $t$ ($1 \geq \rho_i \geq 1-\modelz_i(0)/\modelz_i^*$). We emphasise that the purpose of conditions (i)-(ii) is to make the analysis more concise by limiting the number of possible scenarios to be considered in our analysis of epoch-wise double descent. By similar arguments as used in the following analysis, the results are believed to be extendable beyond these conditions.

Since inactive weights will not contribute to the dynamics of the generalisation error, we focus on active weights, i.e. $\modelz_i(t)$ for which $i \in \activeset$. Under conditions (i)-(ii), the following additional conditions must be fulfilled for a weight $\modelz_i(t)$ to be active: 
\begin{enumerate}[label=(\roman*)]
    \setcounter{enumi}{2}
    \item As additions to condition (i):
    \begin{itemize}
        \item If $\gamma_i = 0$, then $\modelz_i(0) \in (0, \truez_i]$.
        \item If $\truez_i = \modelz_i^*$ ($\rho_i=0$), then $\modelz_i(0) \in [0, \truez_i)$.
    \end{itemize}
    \item The $i^{\tht}$ singular value of $n^{-1}\yvecdd^\top \xvecdd$ is non-zero, i.e.  $\lambdayxi > 0$. 
    \item If $\lr=0$, the learning rates $\lr_a, \lr_b$ and the initialisation of $\modelz_i(t)$ are chosen such that $\gamma_i \neq 0$. Similarly, if $\gamma_i = 0$, then the learning rates $\lr_a, \lr_b$ are chosen such that $\lr > 0$.
\end{enumerate}
Any diagonal weight $\modelz_i(t)$ fulfilling conditions (i)-(ii), but not all of the conditions (iii)-(v) will be inactive, as either the weight is initialised at a fixed point, where it will remain, or the \textit{effective} learning rate $\sqrt{\mykai}$ is $0$. Notice that, since $\rank{\yvecdd^\top \xvecdd} \leq \rank{\xvecdd^\top \xvecdd}$, condition (iv) implies $\lambda_i > 0, \, \forall i \in \activeset$, and that \cref{prop:dyn_sol} therefore holds under \assump.

Using condition (ii) and the reparametrisation of $\truez_i$ in terms of the parameter $\rho_i$, we rewrite a single error term in the sum of \cref{eq:full_test_loss_z} as
\begin{align}
    \lossgeni(t) =  \big((1-\rho_i)\lambdayxi - \lambda_i\mzti \big)^2.
    \label{eq:test_loss_single_curve}
\end{align}
The parameter $\rho_i$ determines the relative position of the true minimum, $\truez_i$, to the initialisation $\modelz_i(0)$ as well as the global minimum $\modelz_i^*$, and hence the form of the generalisation curve. As part of the following analysis, we will consider the time it takes for $\modelz_i(t)$ to reach this true minimum, starting at $\modelz_i(0)$. Under \assump this time, denoted $t_i^{(1-\rho_i)\modelz_i^*}$, can be expressed according to
\begin{align}
     t_i^{(1-\rho_i)\modelz_i^*} = \frac{\log \Bigg(\frac{\sqrt{\Big(\mykai\Big)\Big(\gamma_i^2 \lambda_i^2 + 4\lr^2 {\lambdayxi}^2(1-\rho_i)^2 \Big)}+ \gamma_i^2 \lambda_i^2 + 4  \lr^2 {\lambdayxi} (1-\rho_i) }{C\lambda_i^2\lambdayxi \rho_i}\Bigg)}{\sqrt{\mykai}}, 
     \label{eq:time_to_min}
\end{align}
with the constant $C$ defined in \cref{prop:dyn_sol}. See \cref{app:sec:dyn_sol} for the derivation.

Starting from \cref{eq:test_loss_single_curve}, we provide the following two lemmas characterising the behaviour of the error curve $\lossgeni(t)$.

\begin{lemma}
    Consider the weight $\modelz_i(t)$ following the dynamics in \cref{eq:two_layer_dynamics_z_relaxed}, under \assump. If $1 \geq \rho_i > 0$, the generalisation curve of $\mzti$, \cref{eq:test_loss_single_curve}, is convex in $t$ at $\truez_i$. If instead $\rho_i=0$, the curve is convex leading up to the point $\truez_i$, while $\truez_i$ itself is an undulation point. 
    \label{lemma:convex_minima}
\end{lemma}

\begin{lemma}
    Consider the weight $\modelz_i(t)$ following the dynamics in \cref{eq:two_layer_dynamics_z_relaxed}. Under \assump, the generalisation curve for $\modelz_i(t)$, \cref{eq:test_loss_single_curve}, has a maximum of three inflection points on the interval $[0, \infty)$, whereof a maximum of two lies in the interval $(0, t_i^{(1-\rho_i)\modelz_i^*})$. Moreover, in the case that three inflection points exist and if $5/7 > \rho_i > 0$, then exactly two inflection points lie in the interval $(0, t_i^{(1-\rho_i)\modelz_i^*})$ and one in the interval $(t_i^{(1-\rho_i)\modelz_i^*}, \infty)$. \label{lemma:inflection_points}
\end{lemma}
We leave the proof of \cref{lemma:convex_minima,lemma:inflection_points} to \cref{app:sec:convex_minima,app:sec:infl_points}.
We highlight the following special cases of \cref{lemma:inflection_points}:
\begin{itemize}
    \item  \textbf{One-layer dynamics ($\boldsymbol{\lr=0}$).} For $1 \geq \rho_i > 0$, the error curve has one inflection point at 
    \begin{align}
        \hat{t}_i^{+} =  \frac{\log\left(\displaystyle\frac{2\big(\lambdayxi - \lambda_i \modelz_i(0)\big)}{\lambdayxi \rho_i} \right)}{|\gamma_i| \lambda_i},
        \label{eq:one_layer_infl_point}
    \end{align}
    lying in the interval $(t_i^{(1-\rho_i)\modelz_i^*}, \infty)$.
    For $\rho_i = 0$, the curve has no inflection points. 
    \item \textbf{Balanced two-layer dynamics ($\boldsymbol{\gamma_i=0}$).} For $1 > \rho_i > 0$, the error curve has up to two inflection points, at
    \begin{align}
        \hat{t}_i^{\pm} = \frac{\log \left(\displaystyle\frac{\big(\lambdayxi -\lambda_i \modelz_i(0)\big)\big(1\pm\sqrt{\rho_i^2-\rho_i+1}\big) }{\lambda_i\rho_i \modelz_i(0)}\right)}{2 \lr \lambdayxi}.
        \label{eq:balanced_infl_points}
    \end{align}
     If it exists, the first inflection point lies in the interval  $(0, t_i^{(1-\rho_i)\modelz_i^*})$, while the second inflection point lies interval $(t_i^{(1-\rho_i)\modelz_i^*}, \infty)$. If $\rho_i=0$, the curve has one potential inflection point at
    \begin{align*}
        \hat{t}_i^{-} = \frac{\log \left(\displaystyle\frac{\lambdayxi - \lambda_i \modelz_i(0)}{2\lambda_i \modelz_i(0)}\right)}{2 \lr \lambdayxi},
    \end{align*}
    that, if it exists, lies in the interval $(0, t_i^{(1-\rho_i)\modelz_i^*})$. In other words, for the balanced dynamics, the curve has up to two inflection points for $1 > \rho_i > 0$, lying on either side of $t_i^{(1-\rho_i)\modelz_i^*}$, but only one inflection point if $\rho_i=0$. For $\hat{t}^{-}$ to exist, we require $\modelz_i(\hat{t}^{-}) \in (\modelz_i(0), \truez_i)$. 
\end{itemize}
\Cref{lemma:convex_minima,lemma:inflection_points} tell us something about the general form of each individual error curve $\lossgeni(t),\, i \in \activeset$, as given by \cref{eq:test_loss_single_curve}. We observe that under the one-layer model ($\lr=0$), the error curve $\lossgeni(t)$ has a maximum of one inflection point, with the curve being convex to start, and potentially becoming concave at a time point after $t^{(1-\rho_i)\modelz_i^*}$. On the other hand, the error curve under the decoupled two-layer model can have up to three inflection points. With this, the error curve is potentially concave on parts of the time interval $[0, t^{(1-\rho_i)\modelz_i^*})$. For the balanced dynamics, for example, the curve can, depending on the initialisation, be concave to begin with, consistent with the initial plateau in learning previously observed for two-layer linear networks with small initialisation \parencite{Advani2020}.

We will consistently use $\hat{t}_i^{-}$ to denote the \textit{maximum} inflection point of the error curve $\lossgeni(t)$ located in the interval $(0, t^{(1-\rho_i)\modelz_i^*})$ and $\hat{t}_i^{+}$ the \textit{minimum} inflection point located in the interval $(t^{(1-\rho_i)\modelz_i^*}, \infty)$. In other words, provided that they exist, $\hat{t}_i^{-}$ and $\hat{t}_i^{+}$ are the two inflection points located closest to, and on either side of, the minimum of $\lossgeni(t)$. Following \cref{lemma:convex_minima}, the error curve is convex in between these points. We will use this to understand under which conditions epoch-wise double descent may occur.

\subsubsection{Necessary condition for epoch-wise double descent}

We have analysed the behaviour of the individual error curves of the total generalisation error in \cref{eq:full_test_loss_z}. Following the conclusion that epoch-wise double descent in the decoupled two-layer model emerges in the superposition of such curves, we use our insights to find a necessary condition for epoch-wise double descent. 

\begin{proposition}
    Consider the weight matrix $\Modelz(t)$ with $|\activeset|$ active weights, following \cref{eq:two_layer_dynamics_z_relaxed}. The total generalisation error, \cref{eq:full_test_loss_z}, is a sum of $|\activeset|$ error curves. Under \assump, a necessary condition for this generalisation error to exhibit a double descent pattern over the course of learning, is that we can find at least one inflection point, $\hat{t}$, belonging to either one of the $|\activeset|$ individual error curves, such that 
    \begin{align*}
        \min \{t_i^{(1-\rho_i)\modelz_i^*}; i \in \activeset \} < \hat{t} < \max \{t_i^{(1-\rho_i)\modelz_i^*}; i \in \activeset \}.
    \end{align*}
    \label{prop:double_descent_necessary_general}
\end{proposition}

\Cref{prop:double_descent_necessary_general} gives a necessary, but not sufficient, condition for epoch-wise double descent in the case of a general number of active weights $|\activeset|$. For brevity, we will study this condition further in the case of two active weights, $\modelz_i(t)$ and $\modelz_j(t)$.

 \begin{corollary}
    \textit{ Consider the weight matrix $\Modelz(t)$ with two active weights, $\modelz_i(t)$ and $\modelz_j(t)$, following \cref{eq:two_layer_dynamics_z_relaxed} and for which, without loss of generality, we assume $t_i^{(1-\rho_i)\modelz_i^*} < t_j^{(1-\rho_j)\modelz_j^*}$. The total generalisation error, \cref{eq:full_test_loss_z}, is a sum of two error curves. Under \assump, a necessary condition for this generalisation error to exhibit a double descent pattern over the course of learning, is that we can find at least one inflection point, $\hat{t}$, belonging to either one of the two individual error curves, such that 
    \begin{align*}
        t_i^{(1-\rho_i)\modelz_i^*} < \hat{t} < t_j^{(1-\rho_j)\modelz_j^*}.
    \end{align*}
    Moreover, provided that they exist, let $\hat{t}_j^{-}$ denote the maximum inflection point of the error curve belonging to $\modelz_j(t)$, and lying on the interval $(0, t_j^{(1-\rho_j)\modelz_j^*}$ and let $\hat{t}_i^{+}$ denote the minimum inflection point belonging to the error curve of $\modelz_i(t)$ and lying on the interval $(t_i^{(1-\rho_i)\modelz_i^*}, \infty)$. The necessary condition for epoch-wise double descent simplifies to fulfilling one of the following two conditions
        \begin{align*}
            t_i^{(1-\rho_i)\modelz_i^*} < \hat{t}_j^{-},\\
             \hat{t}_i^{+} < t_j^{(1-\rho_j)\modelz_j^*}.
        \end{align*}
    }    
    \label{prop:double_descent_necessary}
\end{corollary}

From \cref{prop:double_descent_necessary}, we observe that when $\Modelz$ has two active weights, $\modelz_i(t)$ and $\modelz_j(t)$, epoch-wise double descent can happen in one of two scenarios. From an intuitive perspective, we might interpret the first scenario, $t_i^{(1-\rho_i)\modelz_i^*} < \hat{t}_j^{-}$, as one where learning of the first weight, $\modelz_i(t)$, tapers off at a point where the generalisation error of the second weight, $\modelz_j(t)$, is still improving. Perhaps particularly so in the case $5/7 > \rho_i > 0$, where the minimum inflection point $\hat{t}_i^{+}$ is the only inflection point of $\lossgeni(t)$ located after $t_i^{(1-\rho_i)\modelz_i^*}$ (see \cref{lemma:convex_minima}). Similarly, we might interpret the second scenario, $\hat{t}_i^{+} < t_j^{(1-\rho_j)\modelz_j^*}$, as one where learning of the second weight, $\modelz_j(t)$, starts to progress first after the first weight, $\modelz_i(t)$, shows signs of overfitting. We refer to \cref{app:sec:dd_necessary} for the proofs of \cref{prop:double_descent_necessary_general,prop:double_descent_necessary}. 

Interestingly, we find that, while the necessary condition for epoch-wise double descent under the balanced dynamics, and with two active weights, also includes two scenarios in which double descent can occur, we find that the same necessary condition under the one-layer model only includes one of the two scenarios:
\begin{itemize}
    \item  \textbf{One-layer dynamics ($\boldsymbol{\lr=0}$).} For $1 \geq \rho_i > 0$, let $\hat{t}_i$ be the inflection point of the error curve corresponding to the weight $\modelz_i(t)$, given by \cref{eq:one_layer_infl_point}. Then, the necessary condition for epoch-wise double descent with two active weights, given in \cref{prop:double_descent_necessary}, simplifies to 
    \begin{align*}
        \hat{t}_i < t_j^{(1-\rho_j)\modelz_j^*}.
    \end{align*}
    If $\modelz_i(0)=\modelz_j(0)=0$ and $\gamma_i=\gamma_j$, this condition is equivalent to
    \begin{align*}
         \log \left(\frac{2}{\rho_i}\right)\leq \frac{\lambda_i}{\lambda_j} \log \left(\frac{1}{\rho_j}\right).
    \end{align*}
    Thereby, we obtain a necessary condition for epoch-wise double descent under the one-layer model, that depends on the eigenvalues $\lambda_i, \lambda_j$ of the input covariance matrix, but not the singular values $\lambdayxi, {\sigma_j}$. Note that for $\rho_i = 0$, and when $\lr=0$, there is no inflection point $\hat{t}$ fulfilling the necessary condition of \cref{prop:double_descent_necessary}.

    \item \textbf{Balanced two-layer dynamics ($\boldsymbol{\gamma_i=0 \, \forall i \in \activeset}$).} In the case of the balanced two-layer dynamics, let $\hat{t}_j^{-}, \hat{t}_i^{+}$ be the inflection points given by \cref{eq:balanced_infl_points}, for the error curves corresponding to the weights $\modelz_j(t)$ and $\modelz_i(t)$, respectively. The necessary condition for epoch-wise double descent with two active weights, given in \cref{prop:double_descent_necessary}, corresponds to fulfilling at least one of the following two conditions: \\
      
      \noindent (i) For $1 > \rho_j \geq 0$
      \begin{align*}
          t_i^{(1-\rho_i)\modelz_i^*} < \hat{t}_j^{-}, 
      \end{align*}\\
      (ii) For $1 > \rho_i > 0$
      \begin{align*}
          \hat{t}_i^{+} < t_j^{(1-\rho_j)\modelz_j^*}.
      \end{align*}
      Observe that the second condition is not possible for $\rho_i=0$. 
\end{itemize}
First, we notice that epoch-wise double descent with two active weights, can not happen if $\rho_i=0$, where the error curve belonging to $\mzti$ is monotonically decreasing. This is also true in the general case, as when $\rho_i=0$, we have $t_i^{(1-\rho_i)\modelz_i^*}=t_i^{\modelz_i^*} = \infty$. Similarly, double descent can not occur if $\rho_j = 1$, where the error curve belonging to $\modelz_j(t)$ is monotonically increasing, as for $\rho_j=1$, we have $t_j^{(1-\rho_j)\modelz_j^*}=t_j^{0} = 0$.

More importantly, and as mentioned previously, there is only one scenario where epoch-wise double descent can happen in the one-layer model; namely the scenario where learning of the first weight, $\modelz_i(t)$, starts to taper off at a point where the generalisation error of the second weight, $\modelz_j(t)$, is still improving. Meanwhile, the necessary condition for epoch-wise double descent in the case $\gamma_i=0 \, \forall i \in \activeset$, i.e. the balanced dynamics, include the same two scenarios as in the general case, given by \cref{prop:double_descent_necessary}, but where $t_j^{-}, t_i^{+}$ are the only two inflection points that can lie in the interval $(t_i^{(1-\rho_i)\modelz_i^*}, t_j^{(1-\rho_j)\modelz_j^*})$.

Furthermore, for vanishing initialisation, we obtain a necessary condition for epoch-wise double descent in the one-layer model that is independent of the singular values of the input-output covariance matrix. This is in accordance with previous observations where only the eigenvalues $\lambda_i, i \in \activeset$, and not the singular values $\lambdayxi, i \in \activeset$, have been identified as factors of epoch-wise double descent, attributing double descent to input features being learned at different speeds, see e.g., \textcite{Heckel2021,Pezeshki2022,Bodin2022}. 

\begin{figure}
    \begin{subfigure}{\textwidth}
        \includegraphics[width=\textwidth]{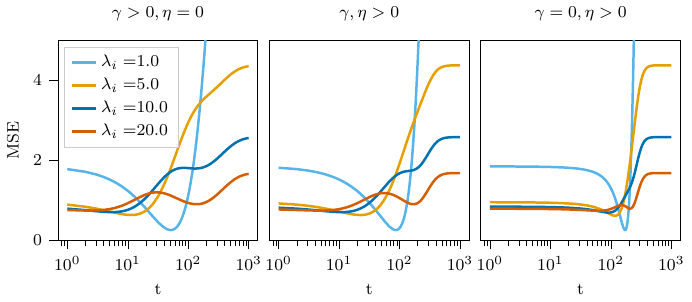}
    \end{subfigure}
    \begin{subfigure}{\textwidth}
        \includegraphics[width=\textwidth]{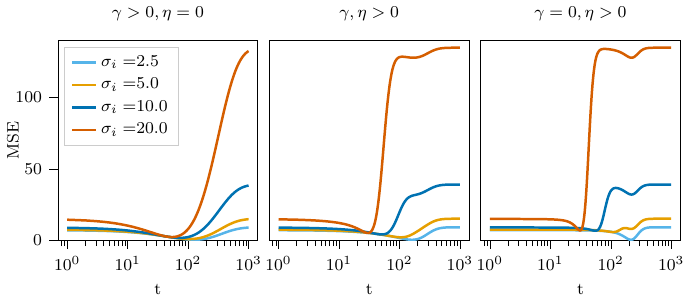}
    \end{subfigure}
    \caption{Examples of double descent with $|\activeset|=10$ active weights, and where active weights are divided into two sets; one set evolving as $\modelz_i(t)$ and the other as $\modelz_j(t)$. We let $\gamma_i=\gamma_j=\gamma$ and consider three scenarios with different values of the parameters $\gamma, \lr$ (Left: one-layer dynamics with $\gamma=0.005, \lr=0$. Middle: bridged dynamics with $\gamma=0.0025, \lr=0.0025$. Right: balanced dynamics with $\gamma=0, \lr=0.005$.). As default, remaining parameters are set according to $\lambda_i = \lambda_j = 1.0, \lambdayxi=\sigma_j = 2.5, \rho_i=0.5, \rho_j=0.8$ and $\modelz_i(0)=\modelz_j(0)= 0.01$. \textbf{Top:} Changing $\lambda_i$. The weight $\modelz_i(t)$ has multiplicity $9$, while $\modelz_j(t)$ has multiplicity $1$. We observe double descent for large $\lambda_i$ in all three scenarios, but double descent seems to appear for a smaller $\lambda_i$ when $\gamma$ is larger. \textbf{Bottom:} Changing $\lambdayxi$. The weight $\modelz_i(t)$ has multiplicity $1$, while $\modelz_j(t)$ has multiplicity $9$. We observe double descent for large $\lambdayxi$ only in the scenarios where $\lr > 0$. }
    \label{fig:mse_changing_lambda_lambdayx}
\end{figure}

In simulations, as shown in \cref{fig:mse_changing_lambda_lambdayx}\footnote{Code for generating the figure is provided at \url{https://github.com/AOlmin/epoch_wise_dd_two_layer_nns}}, 
we also do not observe a relationship between epoch-wise double descent and the singular values $\lambdayxi, i \in \activeset$, for the one-layer model, while we do observe such a dependence for the two-layer model (bottom row of \cref{fig:mse_changing_lambda_lambdayx}). Considering the balanced two-layer model, for which $\gamma_i=0 \, \forall i \in \activeset$, our findings are in accordance with observations made by e.g. \textcite{Gidel2019}, where the singular values $\lambdayxi, i \in \activeset$, determines the learning speed at vanishing initialisation, although these observations have not been connected to epoch-wise double descent. Moreover, the findings align with \cref{prop:dyn_sol}, where the singular values of the input-output covariance matrix have an effect on the convergence rate of the weight $\mzti$ only when $\lr > 0$. Although, here we do find that also the eigenvalues $\lambda_i$, $i \in \activeset$, can impact epoch-wise double descent in the case $\gamma_i=0 \, \forall i \in \activeset$, which is not indicated by this convergence rate.

In practice, we find that for large differences between eigenvalues ($\lambda_i$ and $\lambda_j$) or singular values ($\lambdayxi$ and $\sigma_j$), one error curve tends to dominate the sum of \cref{eq:full_test_loss_z}, and thereby the double descent pattern is concealed. For the simulations in \cref{fig:mse_changing_lambda_lambdayx}, we therefore mimic the scenario where several weights have either large eigenvalues or small singular values, by duplicating one of the weights $\modelz_i(t)$ and $\modelz_j(t)$.  

Apart from the eigenvalues $\lambda_i, i \in \activeset$ as well as the singular values $\lambdayxi, i \in \activeset$, another factor that could potentially cause double descent, are relative distances to the true weights $\truez_i, i \in \activeset$ in the different dimensions, depending on the parameter $\rho_i$. The true noise model will indeed have an effect on double descent also in the single-layer model, see e.g. \textcite{Stephenson2021}. Moreover, differences in initialisation and learning rates, could also be factors affecting the form of the generalisation curve in \cref{eq:full_test_loss_z}, including the potential emergence of a double descent pattern.

\section{Discussion}
We investigate epoch-wise double descent in decoupled two-layer linear networks, identifying factors of epoch-wise double descent in gradient-based learning. First, apart from differences in the eigenvalues of the input covariance matrix of the training data, also differences in the singular values of the input-output covariance matrix can cause epoch-wise double descent in the two-layer model. Second, by providing a necessary condition for epoch-wise double descent both for the standard linear regression model and for the linear decoupled two-layer network, we reveal an additional circumstance under which epoch-wise double descent can emerge in the two-layer model, not observed for the one-layer model. 

There are several venues for further investigating epoch-wise double descent in (linear) neural networks, with the aim to better understand the causes of epoch-wise double descent in deep models. For one, it remains to be investigated if patterns in the training loss, such as the previously observed incremental learning pattern (see e.g \textcite{Gidel2019, Gissin2019,Berthier2023}) could be indicators of epoch-wise double descent. Moreover, several assumptions was made leading up to the dynamics in \cref{eq:two_layer_dynamics_z_relaxed} and relaxing these assumptions could render new insights into the epoch-wise double descent phenomenon. We shortly discuss two extensions, namely that of deep(er) linear neural networks, as well as that of studying the dynamics of \cref{eq:synaptic_dynamics} without decoupling of the weights.

\paragraph{Incremental learning and epoch-wise double descent}
Integral to understanding learning dynamics is the training loss, \cref{eq:full_loss}. Previous studies on two-layer linear neural networks have discovered an implicit bias of gradient descent, causing an incremental learning pattern where weights are learned sequentially and where the training loss typically exhibits plateaus of close to constant error, see e.g., \textcite{Gissin2019,Gidel2019,Berthier2023}. It would be an interesting venue to investigate whether incremental learning, bearing similarities with the epoch-wise double descent phenomenon, could be an indicator of epoch-wise double descent.
\paragraph{Deeper models}

While both the standard linear regression model and the two-layer linear neural network are indeed linear models, we have seen that they exhibit different learning behaviours. Hence, as the number of layers, $L$, grows, we might expect the dynamics to change further. Let us consider the linear neural network with $L$ layers, where the weight matrix in the synaptic weight space can be described as
\begin{align*}
    \Modelz = {\Uyx}^\top \Modelw \V = \prod_{\ell=1}^{L} \Layeriz{\ell},
\end{align*}
each layer $\ell$ represented by the weight matrix $\Layeriz{\ell}$ and with its own number of hidden units. Similar to the two-layer model, we can decouple the weights of $\Modelz(t)$, by initialising this weight matrix to be diagonal, and such that diagonal weights evolve independently. Then, to obtain an approximate dynamics for the $i^{\tht}$ diagonal element, $\modelz_i(t)$, of $\Modelz(t)$, we make a number of simplifications. These simplifications include assuming $L$ to be even and large, as well as grouping the model layers into two equally sized groups, where layers within the same group have the same initialisation and learning rate. We denote the learning rates of the two groups by $\lr_a$ and $\lr_b$, respectively. The approximate dynamics are
\begin{align}
     \frac{d\modelz_i}{dt} \approx \frac{L}{2} \sqrt{\gamma_i + 4 \lr^2}\modelz_i^2 (\lambdayxi - \lambda_i \modelz_i), 
     \label{eq:deeper_models_dynamics_z}
\end{align}
where we (again) define $\lr \coloneqq \sqrt{\lr_a \lr_b}$. The parameter $\gamma_i$ is a conservation quantity, analogous to the same parameter in the decoupled two-layer dynamics in \cref{eq:two_layer_dynamics_z_relaxed}. The approximate dynamics of the multi-layer linear neural network are similar to the dynamics found by \textcite{Saxe2014}, but differ in two regards. First, while \textcite{Saxe2014} assume that $\gamma_i$=0, this parameter has a direct influence on the \textit{effective} learning rate $\sqrt{\gamma_i + 4 \lr^2}$ in \cref{eq:deeper_models_dynamics_z}. Second, in our dynamics, $\lambda_i$ is possibly different from $1$. We refer to \cref{app:sec:deeper_models} for details on the derivation. 

We observe that the dynamics in \cref{eq:deeper_models_dynamics_z} are cubic in $\modelz_i(t)$. However, with the simplifying assumptions made here, we can rewrite the total generalisation error of the multi-layer model as a sum over error curves corresponding to active weights $\mzti$, for $i \in \activeset$, as in \cref{eq:full_test_loss_z}. Moreover, under \assump, the error curve $\lossgeni(t)$ corresponding to an active weight $\mzti$, and as given by \cref{eq:test_loss_single_curve}, will be monotonically decreasing, U-shaped, or monotonically increasing in $t$. We can also show that such an error curve will have at most two inflection points (see \cref{app:sec:deeper_models}). Taken together, this indicates a necessary condition for epoch-wise double descent in the multi-layer model, that is similar to the one given by \cref{prop:double_descent_necessary_general}. Nevertheless, with more relaxed assumptions, the error curve for the multi-layer dynamics might show more complex behaviour, potentially with additional, unidentified factors causing epoch-wise double descent in deeper models. 

\paragraph{The effect of coupling}
In the decoupled dynamics, \cref{eq:two_layer_dynamics_z_relaxed}, the weights in both the first and second layer of the two-layer network are each connected only to a single set of eigenvalue $\lambda_i$ and singular value $\lambdayxi$. If dynamics are not decoupled, however, weights can be connected to several eigenvalues of the input covariance matrix as well as several singular values of the input-output covariance matrix, resulting in interaction between weights in the learning dynamics, see e.g., \textcite{Saxe2014,Advani2020}. We might expect such an interaction to give rise to new types of epoch-wise double descent behaviours. For example, \textcite{Heckel2021} hypothesise that epoch-wise double descent in two-layer neural networks with ReLU activation emerges because the weights in the second layer are learnt faster than the first layer weights. This could indicate a trade-off between independent evolution of weights and evolution through interaction. While not observed for the decoupled two-layer dynamics, it remains to be investigated if such a behaviour could also be observed in the (coupled) linear two-layer neural network.

\subsubsection*{Acknowledgements}
This research is financially supported by the Swedish Research Council, the Wallenberg AI, Autonomous Systems and Software Program (WASP) funded by the Knut and Alice Wallenberg Foundation,
and
the Excellence Center at Linköping--Lund in Information Technology (ELLIIT).

\newpage
\bibliographystyle{abbrvnat} 
\bibliography{references}

\newpage
\appendix

\noindent\LARGE{\textbf{Supplementary material for \\ \textit{Towards understanding epoch-wise double descent in two-layer linear neural networks}}}
\normalsize
\section{Theoretical derivations}
We provide derivations of the decoupled two-layer linear neural network dynamics as well as proofs for the main results of the paper. In the following derivations, we will use $A_{i, :}$ and $A_{:, i}$ to denote the $i^{\tht}$ row and column, respectively, of any matrix $A$.

\subsection{Decoupled dynamics of two-layer linear networks}
\label{app:sec:decoupled_dynamics}
To derive the decoupled dynamics in \cref{eq:two_layer_dynamics_a_b_relaxed}, we start by deriving the dynamics of the weights $\Layeri{1}$ and $\Layeri{2}$, using gradient descent and the MSE criterion given by \cref{eq:full_loss}.
The derivatives of the MSE criterion are
\begin{align*}
    &\frac{d\losstrain}{d\Modelw} = \frac{1}{n}(
 \yvecdd^\top \xvecdd - \Modelw \xvecdd^\top \xvecdd) = ( U^{(yx)} \Sigma {V^{(yx)}}^\top - \Modelw V\Lambda V^\top),\\
    &\frac{d\losstrain}{d\Layeri{1}} = {\Layeri{2}}^\top \frac{d\losstrain}{d\Modelw},  \\
     & \frac{d\losstrain}{d\Layeri{2}} = \frac{d\losstrain}{d\Modelw} {\Layeri{1}}^\top,
\end{align*}
where we use $ \n^{-1}  \yvecdd^\top \xvecdd =  U^{(yx)} \Sigma {V^{(yx)}}^\top$ and $\n^{-1}\xvecdd^\top \xvecdd = V\Lambda V^\top$. This yields the dynamics
\begin{align*}
    &   \frac{1}{\lr_a}\frac{d}{dt} \Layeri{1} = {\Layeri{2}}^\top( U^{(yx)} \Sigma {V^{(yx)}}^\top - \Modelw V\Lambda V^\top),  \\
    &    \frac{1}{\lr_b}\frac{d}{dt} \Layeri{2} = ( U^{(yx)} \Sigma {V^{(yx)}}^\top - \Modelw V\Lambda V^\top) {\Layeri{1}}^\top,
\end{align*}
with learning rates $\lr_a, \lr_b \geq 0$. Note that we do not yet assume $V = V^{(yx)}$, as is assumed in the main paper. We will introduce this assumption at a later stage in the derivations.

Following \textcite{Saxe2014}, we introduce the \textit{synaptic} weights $\Modelz \coloneqq \Layeriz{2}\Layeriz{1}$ with $\Layeriz{1} \coloneqq \Layeri{1} {V^{(yx)}}$, $\Layeriz{2} \coloneqq {U^{(yx)}}^\top\Layeri{2}$. The dynamics for $\Layeriz{1}$ are
\begin{align*}
    \frac{1}{\lr_a}\frac{d}{dt} \Layeriz{1} &= \frac{1}{\lr_a}\frac{d}{dt} \Layeri{1} V^{(yx)} \\ &= {\Layeri{2}}^\top( U^{(yx)} \Sigma {V^{(yx)}}^\top - \Modelw V\Lambda V^\top) V^{(yx)} \\ &=
    {\Layeri{2}}^\top( U^{(yx)} \Sigma - U^{(yx)} \Modelz {V^{(yx)}}^\top V\Lambda V^\top V^{(yx)}) \\ &= 
    {\Layeriz{2}}^\top( \Sigma  - \Modelz {V^{(yx)}}^\top V\Lambda V^\top V^{(yx)}). 
\end{align*}
Similarly, the dynamics for $\Layeriz{2}$ are
\begin{align*}
    \frac{1}{\lr_b}\frac{d}{dt} \Layeriz{2} &= \frac{1}{\lr_b} {U^{(yx)}}^\top \frac{d}{dt} \Layeri{2} \\ &= {U^{(yx)}}^\top (U^{(yx)} \Sigma {V^{(yx)}}^\top - \Modelw V\Lambda V^\top)  {\Layeri{1}}^\top \\ &=
    (\Sigma - \Modelz {V^{(yx)}}^\top V\Lambda V^\top V^{(yx)}) {\Layeriz{1}}^\top.
\end{align*}
Let $V = \big[V_{:, \nu_1}^{(yx)}, V_{:, \nu_2}^{(yx)}, \ldots V_{:, \nu_{\xdim}}^{(yx)}\big]$ for indices $\nu_1, \nu_2, \ldots \nu_{\xdim} \in \{1, 2, \ldots, \xdim\}$, with $\nu_i \neq \nu_j$ for $i \neq j$. In other words, we assume that the unitary matrix $V$ is equal to $V^{(yx)}$ but with permuted columns. Then, the $(i, j)^\tht$ element of the matrix $V^\top {V^{(yx)}}$ is
\begin{align*}
    (V^\top {V^{(yx)}})_{i, j} = V_{:, \nu_i}^{(yx)} \cdot V_{:, j}^{(yx)} = \begin{cases}
        1, \quad &\text{if}\; \nu_i = j,\\
        0, & \text{otherwise},
    \end{cases}
\end{align*}
with ''$\cdot$'' the dot product. Subsequently, the dynamics of $\Layeriz{1}$ and $\Layeriz{2}$ simplifies to 
\begin{align*}
    \frac{1}{\lr_a}\frac{d}{dt} \Layeriz{1} &= 
    {\Layeriz{2}}^\top(\Sigma  -  \Modelz \tilde{ \Lambda}), \\ 
    \frac{1}{\lr_b}\frac{d}{dt} \Layeriz{2} &=
    (\Sigma -  \Modelz \tilde{ \Lambda} ) {\Layeriz{1}}^\top,
\end{align*}
where the diagonal matrix $\tilde{\Lambda}$ is equal to $\Lambda$ but with diagonal elements reshuffled, such that $\tilde{\Lambda}_{i, i} = \Lambda_{\nu_i, \nu_i}$.

Now, let $\alpha^{(i)}$ denote the $i^{\tht}$ column of $\Layeriz{1}$ and ${\beta^{(i)}}^\top$ denote the $i^{\tht}$ row of $\Layeriz{2}$. Moreover, let  $\tilde{\lambda}_i$ and $\lambdayxi$ be the $i^{th}$ diagonal elements of $\tilde{\Lambda}$ and $\Sigma$, respectively. We derive the dynamics of $\alpha^{(i)}$, for $i = 1, \ldots, \xdim$, and $\beta^{(i)}$, for $i = 1, \ldots, \ydim$.

First, we note that the synaptic weight matrix $\Modelz$ has elements
\begin{align*}
    \Modelz_{i, j} = \alpha^{(j)} \cdot \beta^{(i)},
\end{align*}
and therefore
\begin{align*}
    (\Modelz \tilde{\Lambda})_{i, j} = (\alpha^{(j)} \cdot \beta^{(i)}) \tilde{\lambda}_{j}.
\end{align*}
Let $S \coloneqq \Sigma -  \Modelz \tilde{\Lambda}$, with
\begin{align*}
    S_{i, j} = \begin{cases}
        \lambdayxi - (\alpha^{(i)} \cdot \beta^{(i)}) \tilde{\lambda}_{i}, \quad & \text{if } j=i,\\
        - (\alpha^{(j)} \cdot \beta^{(i)}) \tilde{\lambda}_{j}, & \text{otherwise}.
    \end{cases}
\end{align*}
Then, for finding the dynamics of the column $\alpha^{(i)}$, we have
\begin{align*}
    ({\Layeriz{2}}^\top S)_{i, j} = \sum_{k=1}^{\ydim} \beta_i^{(k)} S_{k, j},
\end{align*}
and, hence,
\begin{align*}
    \frac{1}{\lr_a}\frac{d}{dt}\alpha^{(i)} &= \sum_{j} \beta^{(j)} S_{j, i} \\ & = \begin{cases}
    (\lambdayxi - \tilde{\lambda}_{i}(\alpha^{(i)} \cdot \beta^{(i)}) ) \beta^{(i)} - \tilde{\lambda}_{i}\sum_{j \neq i} (\alpha^{(i)} \cdot \beta^{(j)}) \beta^{(j)},  \quad & \text{if } i \leq \ydim,\\
    - \tilde{\lambda}_{i}\sum_{j} (\alpha^{(i)} \cdot \beta^{(j)}) \beta^{(j)}, & \text{otherwise}.
    \end{cases}
\end{align*}

For the dynamics of the row $\beta^{(i)}$, note that
\begin{align*}
    ( S {\Layeriz{1}}^\top)_{i, j} = \sum_{k=1}^{\xdim} S_{i, k} \alpha_j^{(k)}.
\end{align*}
Therefore, 
\begin{align*}
    \frac{1}{\lr_b}\frac{d}{dt}\beta^{(i)} &= \sum_{j} S_{i, j} \alpha^{(j)} \\ &=  \begin{cases}
        (\lambdayxi - \tilde{\lambda}_{i}(\alpha^{(i)} \cdot \beta^{(i)}) ) \alpha^{(i)} - \sum_{j \neq i} \tilde{\lambda}_j (\alpha^{(j)} \cdot \beta^{(i)}) \alpha^{(j)}, \quad & \text{if } i \leq \xdim,\\
        - \sum_{j} \tilde{\lambda}_j (\alpha^{(j)} \cdot \beta^{(i)}) \alpha^{(j)}, & \text{otherwise}.
    \end{cases}
\end{align*}
We decouple weights by initialising $\alpha^{(i)}, {\beta^{(i)}} \, \propto \, r^{(i)}$, for $i=1, \ldots, \min(\xdim, \ydim)$. The (constant) vectors $r^{(i)}$, for $i=1, \ldots, \min(\xdim, \ydim)$, are orthogonal unit vectors, such that $r^{(i)} \cdot r^{(i)} = 1$ and $r^{(i)} \cdot r^{(j)} = 0$ if $i \neq j$. Note that decoupling also entails initialising any columns or rows with index larger than $\min(\xdim, \ydim)$ to the zero vector. Following this initialisation, we obtain the decoupled dynamics 
\begin{align*}
    \frac{1}{\lr_a}\frac{d}{dt}\alpha^{(i)} &= 
    (\lambdayxi - \tilde{\lambda}_{i}(\alpha^{(i)} \cdot \beta^{(i)}) ) \beta^{(i)} \\
    \frac{1}{\lr_b}\frac{d}{dt}\beta^{(i)} &=  (\lambdayxi - \tilde{\lambda}_{i}(\alpha^{(i)} \cdot \beta^{(i)}) ) \alpha^{(i)} 
\end{align*}
Once decoupled, weights remain decoupled, as for any $i$ and $j$, with $j \neq i$, we have
\begin{align*}
    \frac{d}{dt} (\alpha^{(i)} \cdot \beta^{(j)}) & =   \lr_a(\lambdayxi - \tilde{\lambda}_{i}(\alpha^{(i)} \cdot \beta^{(i)}) ) \beta^{(i)} \cdot \beta^{(j)} \\ & \eqspace  + \lr_b(\sigma_j - \tilde{\lambda}_{j}(\alpha^{(j)} \cdot \beta^{(j)}) ) \alpha^{(j)} \cdot \alpha^{(i)} = 0,
\end{align*}
following that $\alpha^{(j)} \cdot \alpha^{(i)} = \beta^{(i)} \cdot \beta^{(j)} = 0$.

We recover \cref{eq:two_layer_dynamics_a_b_relaxed}, by rewriting the decoupled dynamics in terms of the scalar projections $a_i = \alpha^{(i)} \cdot r^{(i)}$ and $b_i = \beta^{(i)} \cdot r^{(i)}$, arriving at
\begin{align*}
    \frac{1}{\lr_a}\frac{da_i}{dt} =  \frac{1}{\lr_a}\frac{d}{dt}(\alpha^{(i)} \cdot r^{(i)}) = b_i(\lambdayxi - \tilde{\lambda}_i a_i b_i), \\
     \frac{1}{\lr_b}\frac{db_i}{dt} =  \frac{1}{\lr_b}\frac{d}{dt}(\beta^{(i)} \cdot r^{(i)}) = a_i(\lambdayxi - \tilde{\lambda}_i a_i b_i).
\end{align*}
In addition, in the main paper, we assume $V = V^{(yx)}$ such that $\tilde{\Lambda}=\Lambda$. Then, replace  $\tilde{\lambda}_i$ with $\lambda_i$, denoting the $i^{\tht}$ diagonal element of $\Lambda$.

\subsection{Deriving the bridged dynamics}
\label{app:sec:bridged_dynamics}
We derive the bridged dynamics in \cref{eq:two_layer_dynamics_z_relaxed} from \cref{eq:two_layer_dynamics_a_b_relaxed}. In the following, as we only consider a single weight, we will drop all sub- and superscripts $i$. Then, starting from \cref{eq:two_layer_dynamics_z_relaxed}, let $\modelz = \alpha \cdot \beta = ab$, for which
\begin{align*}
    \frac{dz}{dt} = \frac{d}{dt} ab  = (\lr_b a^2 + \lr_a b^2)(\lambdayx - \lambda \modelz) .
\end{align*}
We note that
\begin{align*}
    \frac{d}{dt}(\lr_b a^2 - \lr_a b^2) = 2 \lr_b a  \frac{da}{dt} - 2 \lr_a b  \frac{db}{dt} = 0.
\end{align*}
Hence, $\lr_b a^2 - \lr_a b^2 \eqqcolon \gamma$ is constant. Subsequently, we can rewrite
\begin{align*}
    \frac{dz}{dt} = (\gamma + 2\lr_a b^2)(\lambdayx - \lambda z).
\end{align*}
Using the conserved quantity $\gamma$, we also have the equality
\begin{align*}
    z = ab = \pm \sqrt{\frac{(\gamma + \lr_a b^2)}{\lr_b}} b,
\end{align*}
from which we find
\begin{align*}
    b = \pm \sqrt{\frac{-\gamma \pm \sqrt{\gamma^2 + 4 \lr_a \lr_b \modelz^2}}{2\lr_a}}.
\end{align*}
Note that only the plus sign in the square root gives a valid solution for general $\lr_a, \lr_b \geq 0, \gamma$. Hence, we have
\begin{align*}
    b^2 = \frac{-\gamma + \sqrt{\gamma^2 + 4 \lr_a \lr_b \modelz^2}}{2\lr_a}.
\end{align*}
Replacing $b^2$ in the current dynamics of $\modelz$ with this expression, we obtain 
\begin{align}
    \frac{dz}{dt} = \sqrt{\gamma^2 + 4 \lr^2 z^2}(\lambdayx - \lambda z),
    \label{eq:supp:bridged_dynamics}
\end{align}
with $\lr = \sqrt{\lr_a \lr_b}$. 

\subsection{Proof of \cref{prop:dyn_sol}}
\label{app:sec:dyn_sol}
For solving the differential equation in \cref{eq:two_layer_dynamics_z_relaxed}, we focus on the case $\modelz(0) < \lambdayx / \lambda$ with $\lambda > 0$. Note that \cref{eq:two_layer_dynamics_z_relaxed} is separable in $z$ and $t$, and so 
\begin{align*}
    \int dt =  \int \frac{dz}{ \sqrt{\gamma^2 + 4 \lr \modelz^2}(\lambdayx - \lambda \modelz)},
\end{align*}
again dropping subscripts $i$ for notational clarity.
Assuming $\lambda > 0$, we solve the integrals to find
\begin{align*}
    t = \frac{\log \left(\displaystyle\frac{\sqrt{\Big(\myka\Big)\Big(\gamma^2 + 4 \lr^2 \modelz^2 \Big)}+ \gamma^2 \lambda + 4  \lr^2 \lambdayx \modelz}{\lambda (\lambdayx - \lambda \modelz)}\right)}{\sqrt{\myka}} + C',
\end{align*}
for a constant $C'$. Note that we from here recover \cref{eq:time_to_min}, by plugging in $z=\truez=(1-\rho)\starz$ and  %
\begin{align*}
    C' = - \frac{\log(C)}{\sqrt{\myka}}.
\end{align*}
Now, solving for $\modelz=\modelz(t)$, with initial value $\modelz(0)$, we obtain
\begin{align*}
    \modelz(t) = \frac{C^2 \lambda^2 \lambdayx e^{2\sqrt{ \myka } t}-2 C \gamma^2 \lambda^2 e^{\sqrt{ \myka } t}-4 \gamma^2 \lr^2 \lambdayx}{\lambda \left(C^2 \lambda^2 e^{2\sqrt{ \myka } t} + 8 C  \lr^2 \lambdayx e^{\sqrt{ \myka } t}-4 \gamma^2 \lr^2\right)},
\end{align*}
with
\begin{align*}
    C = \frac{\sqrt{\big(\myka\big)\big(\gamma^2 + 4 \lr^2 \modelz(0)^2 \big)}+ \gamma^2 \lambda + 4  \lr^2 \lambdayx \modelz(0)}{\lambda (\lambdayx - \lambda \modelz(0))}.
\end{align*}
Note that arriving at this solution, includes solving the equation
\begin{align*}
    &\sqrt{\big(\myka\big)\big(\gamma^2 + 4 \lr^2 \modelz(0)^2 \big)} \\ & \eqspace \eqspace = \lambda \big(\lambdayx - \lambda \modelz(t))\big) C e^{\sqrt{\myka} t} - \gamma^2\lambda - 4\lr^2 \lambdayx \modelz(t),
\end{align*}
for which we require 
\begin{align*}
    \lambda \big(\lambdayx - \lambda \modelz(t))\big) C e^{\sqrt{\myka} t} - \gamma^2\lambda - 4\lr^2 \lambdayx \modelz(t) \geq 0.
\end{align*}
To verify if (and when) this holds, we rewrite the expression in terms of the solution $\modelz(t)$, resulting in
\begin{align*}
    & \lambda \big(\lambdayx - \lambda \modelz(t))\big) C e^{\sqrt{\myka} t} - \gamma^2\lambda - 4\lr^2 \lambdayx \modelz(t) \\ & \eqspace \eqspace = \frac{\big( \myka \big) \big(C^2 \lambda^2 e^{2 \sqrt{\myka} t} + 4 \gamma^2 \lr^2\big)}{\lambda\big(C^2 \lambda^2 e^{2 \sqrt{\myka} t} + 8 C \lr^2 \lambdayx e^{\sqrt{\myka} t} - 4 \gamma^2 \lr^2 \big)}.
\end{align*}
We observe that the nominator is positive. For the denominator, note that if $\modelz(0) < \lambdayx / \lambda$, then $C \geq 0$ and, therefore,
\begin{align*}
   & \lambda\Big(C^2 \lambda^2 e^{2 \sqrt{\myka} t} + 8 C \lr^2 \lambdayx e^{\sqrt{\myka} t} - 4 \gamma^2 \lr^2 \Big) \\ & \eqspace \eqspace \eqspace \eqspace \eqspace   \geq \lambda\big(C^2 \lambda^2  + 8 C \lr^2 \lambdayx- 4 \gamma^2 \lr^2 \big).
\end{align*}
Plugging in the expression for $C$, we find
\begin{align*}
     & \lambda\big(C^2 \lambda^2  + 8 C \lr^2 \lambdayx- 4 \gamma^2 \lr^2 \big)
    \\ & \quad  = \frac{2 \big(\myka\big)}{(\lambdayx - \lambda \modelz(0))^2}  \Big(\sqrt{\big(\myka\big)\big(\gamma^2 + 4 \lr^2 \modelz(0)^2 \big)}  + \gamma^2 \lambda + 4 \lr^2 \lambdayx \modelz(0) \Big). 
\end{align*}
If $\gamma \neq 0$, and following the assumption that $\lambda > 0$, we conclude that this expression is positive. If $\gamma = 0$, we require $\lr > 0$ and $\modelz(0) > 0$ for the expression to be positive (otherwise it equals $0$). 

Hence, the solution holds for  $\modelz(0) < \lambdayx / \lambda$ with $\lambda > 0$ if $\gamma \neq 0$ or, when $\gamma=0$, if in addition $\lr > 0$ and $\modelz(0) > 0$. This finishes the proof of the first part of \cref{prop:dyn_sol}. Before moving on to the next part of the proof, we point out that we recover \cref{eq:one_layer_solution_z,eq:two_layer_solution_z}, by inserting $\lr=0$ and $\gamma=0$, respectively, into the solution.

For the next part of \cref{prop:dyn_sol}, we take the limit of the solution, $\modelz(t)$, to find
\begin{align*}
    \lim_{t \rightarrow \infty} \modelz(t) = \begin{cases}
        \frac{\lambdayx}{\lambda}, \quad &\text{if } \gamma \neq 0, \\
        \frac{\lambdayx}{\lambda}, \quad &\text{if } \gamma = 0, \; \lr, \modelz(0) > 0, \\
        \modelz(0), \quad &\text{if } \gamma = \lr = 0, \\
        0, &\text{otherwise}. 
    \end{cases}
\end{align*}
In addition, observe that, for large $t$, and assuming $\lr, \modelz(0)>0$ if $\gamma = 0$, we have 
\begin{align*}
    \big| \starz - \modelz(t) \big| &= \left| \frac{2 C e^{\sqrt{\myka} t} \left(\gamma^2 \lambda^2 +4\lr^2 \lambdayx^2 \right)}{\lambda \left(C^2 \lambda^2 e^{2 \sqrt{\myka} t}+8 C \lr^2 \lambdayx e^{\sqrt{\myka} t}-4 \gamma^2 \lr^2 \right)} \right| \\ & \simeq \bigg|2C^{-1}\lambda^{-3}( \myka) e^{-\sqrt{\myka} t} \bigg|,
\end{align*}
with $|\cdot|$ denoting an absolute value and with $\starz = \lambdayx / \lambda$.
\subsection{Proof of \cref{lemma:true_synaptic_weight_expr,prop:test_mse}}
\label{app:sec:test_mse}

We derive \cref{eq:full_test_loss_z}, starting from
\begin{align*}
   \lossgen(t) = \frac{1}{2} \expectation_{\x, \y} \Big[ (y - \x \Modelw (t)^\top) (y - \x \Modelw (t)^\top)^\top \Big],
\end{align*}
with  $x \sim \mathcal{N}(\mathbf{0}, \bar{\Lambda})$ and $y$ defined in \cref{eq:true_model}.

First, we prove \cref{lemma:true_synaptic_weight_expr}, by rewriting the true weight matrix $\Truew$ (\cref{eq:true_model}) in terms of the global minimum. We have
\begin{align*}
    \yvecdd^\top \xvecdd (\xvecdd^\top \xvecdd)^{\dagger} = \Uyx \Syx \Lambda^{\dagger} V^\top, 
\end{align*}
where $\dagger$ denotes the Moore-Penrose pseudoinverse. Observe that the global minimum can also be expressed using \cref{eq:true_model} according to
\begin{align*}
    \yvecdd^\top \xvecdd (\xvecdd^\top \xvecdd)^{\dagger} &= (\Truew \xvecdd^\top \xvecdd + \epsvec^\top \xvecdd )  (\xvecdd^\top \xvecdd)^{\dagger} \\ &= (\Truew \V \Lambda  + \n^{-1/2}\epsvec^\top \U \Lambda^{1/2}  )  \Lambda^{\dagger} \V^\top,
\end{align*}
with 
\begin{align*}
    \epsvec = \begin{bmatrix}
        - \, \epsilon_1  \, - \\
        - \, \epsilon_2  \, - \\
        \cdots \\
        - \, \epsilon_n \, - 
    \end{bmatrix},
\end{align*}
the matrix of residual vectors $\epsilon_i = \y_i - \x_i \Truew^\top$, for $i = 1, \ldots, n$. 
Using the two equations for the global minimum, we find
\begin{align*}
    \Truew &= \Uyx (\Syx - \n^{-1/2}{U^{(yx)}}^\top\epsvec^\top U {\Lambda^{1/2}}) \Lambda^{\dagger} V^\top \\ &= \Uyx (\Syx \Lambda^{\dagger} - \widetilde{\epsvec}^\top ({\Lambda^{1/2}})^{\dagger}) \V^\top,
\end{align*}
with $\widetilde{\epsvec} \coloneqq \n^{-1/2} U^\top \epsvec {U^{(yx)}}$. 
Subsequently, following the relation $\Truew = {\Uyx} \Truez {\Vyx}^\top$, the true weights in the synaptic weight space are
\begin{align*}
    \Truez \coloneqq {U^{(yx)}}^\top \Truew \V = \Syx \Lambda^{\dagger} - \widetilde{\epsvec}^\top ({\Lambda^{1/2}})^{\dagger}.
\end{align*} 
Letting $\widetilde{\epsilon}_i$ denote the $i^{\tht}$ element on the main diagonal of $\widetilde{\epsvec}$ and assuming $\lambda_i > 0$, the $i^{th}$ diagonal element of $\Truez$ follows directly as
\begin{align*}
    \truez_i = \lambdayxi \lambda_i^{-1} - \tilde{\epsilon}_i \lambda_i^{-1/2}.
\end{align*}
This finishes the proof of \cref{lemma:true_synaptic_weight_expr}.

Now, in the synaptic weight space, using $x \sim \mathcal{N}(\mathbf{0}, \bar{\Lambda})$ and with $y$ following \cref{eq:true_model}, we have
\begin{align*}
  \lossgen(t) &= \frac{1}{2}\expectation_{\x, \y} \bigg[ \big(\y - \x ({\Uyx}\Modelz(t)\V^\top)^\top\big) \big(\y - \x ({\Uyx}\Modelz(t)\V^\top)^\top \big)^\top \bigg]\\ &= 
    \frac{1}{2}\expectation_{\x, \epsilon} \bigg[ \big(\x \V (\Truez-\Modelz(t))^\top{\Uyx}^\top  + \epsilon \big) \big(\x \V (\Truez-\Modelz(t))^\top{\Uyx}^\top  + \epsilon \big)^\top \bigg]\\ &
    = \frac{1}{2}\expectation_{\x, \epsilon} \bigg[\x \V \big (\Truez - \Modelz(t) \big)^\top  \big (\Truez - \Modelz(t) \big) \V^\top \x^\top \\ & \eqspace\eqspace\eqspace\eqspace  +  2 \x \V (\Truez-\Modelz(t))^\top{\Uyx}^\top \epsilon^\top + \epsilon \epsilon^\top \bigg]  \\ & \overset{(1)}{=} \frac{1}{2}\bigg(\expectation_{\x, \epsilon} \bigg[\x \V \big (\Truez - \Modelz(t) \big)^\top  \big (\Truez - \Modelz(t) \big) \V^\top \x^\top\bigg]  + \trace{\Lambda^{(\epsilon)}}  \bigg)  \\ &
    \overset{(2)}{=} \frac{1}{2} \Big( \text{Tr}\Big(\big (\Truez - \Modelz(t) \big)^\top  \big (\Truez - \Modelz(t) \big) \V^\top \bar{\Sx} \V \Big) + \trace{\Lambda^{(\epsilon)}} \Big) \\ & \overset{(3)}{\approx} \frac{1}{2} \Big(\trace{\big (\Truez - \Modelz(t) \big)^\top  \big (\Truez - \Modelz(t) \big) \Lambda} +  \trace{\Lambda^{(\epsilon)}} \Big) \\ & =  \frac{1}{2} \Big(\text{Tr}\Big(\big(\Syx - \widetilde{\epsvec}^\top ({\Lambda^{1/2}})^{\dagger} - \Modelz(t)\big)^\top \\ &  \eqspace\eqspace \eqspace \eqspace \big(\Syx - \widetilde{\epsvec}^\top ({\Lambda^{1/2}})^{\dagger} - \Modelz(t) \big) \Lambda \Big)  + \trace{\Lambda^{(\epsilon)}} \Big) \\ & = \frac{1}{2}\sum_{i \in \activeset} \lambda_i \big(\lambdayxi \lambda_i^{-1} - \tilde{\epsilon}_i \lambda_i^{-1/2} - z_i(t) \big)^2  + \text{const.}, 
\end{align*}
resulting in \cref{eq:full_test_loss_z}. In (1), the middle term evaluates to $0$, since $\x$ and $\epsilon$ are independent, zero-mean variables. Moreover, in (2), we use $\expectation_{\x}[x A x^\top] = \trace{A\bar{\Lambda}}$, with $A\coloneqq\V \big (\Truez - \Modelz(t) \big)^\top  \big (\Truez - \Modelz(t) \big) \V^\top$, together with the cyclic property of the trace, while we in (3), use the approximation $V^\top \bar{\Lambda} V \approx \Lambda$. The final constant term includes the noise term $\trace{\Lambda^{(\epsilon)}}$ as well as a possible constant error corresponding to constant entries in $Z(t)$, including non-diagonal elements for which corresponding entries in $\Truez$ are non-zero.

For the second part of the proof, consider an individual error curve in the sum of \cref{eq:full_test_loss_z}, which has the form
\begin{align*}
  \lossgeni(t) = \lambda_i \big(\lambdayxi \lambda_i^{-1} - \tilde{\epsilon}_i \lambda_i^{-1/2} - z_i(t) \big)^2,
\end{align*}
with $i \in \activeset$. To show that this error curve is either monotonically decreasing, U-shaped, or monotonically increasing in $t$, we study the gradient
\begin{align*}
   & \frac{d}{dt}\lossgeni(t) = - 2\lambda_i \big(\lambdayxi \lambda_i^{-1} - \tilde{\epsilon}_i \lambda_i^{-1/2} - z_i(t)\big) \frac{d\modelz_i(t)}{dt}.
\end{align*}
Following the dynamics in \cref{eq:two_layer_dynamics_z_relaxed}, the sign of $d\modelz_i(t)/dt$ will be determined by the term $(\lambdayxi-\lambda \modelz_i(t) )$ and, subsequently, on the initialisation. If $\modelz_i(0) \leq \modelz_i^* (=\lambdayxi/\lambda_i)$, $\mzti$ will monotonically increase in $t$ with 
\begin{align*}
    \frac{d\modelz_i(t)}{dt} \geq 0, \; \forall t.
\end{align*}
On the other hand, if  $\modelz_i(0) > \modelz_i^*$, $\mzti$ will monotonically decrease in $t$ with 
\begin{align*}
    \frac{d\modelz_i(t)}{dt} \leq 0, \; \forall t.
\end{align*}
With this in mind, observe that the gradient of the error curve $\lossgeni(t)$ will change signs only at one point. We see this by solving the equation 
\begin{align*}
    \frac{d}{dt}\lossgeni(t) = 0.
\end{align*}
While one solution to this equation is $d\modelz_i(t)/dt = 0$, we note that the gradient will not change signs at this point. Instead, if $d\modelz_i(t)/dt = 0$, $\mzti$ will remain at its current value, and so will $\lossgeni(t)$. The other solution is
\begin{align*}
    \mzti = \lambdayxi \lambda_i^{-1} - \tilde{\epsilon}_i \lambda_i^{-1/2},
\end{align*}
and as $\mzti$ is either monotonically increasing or decreasing in $t$, this point will be passed only once during the course of learning. 

Now, following \cref{eq:two_layer_dynamics_z_relaxed}, $\modelz_i(t)$ will either increase or decrease in $t$, depending on the initialisation, starting at $\modelz_i(0)$ and ending at $\modelz_i^*$. An exception can be seen in the special case $\gamma_i=0$ and $\modelz_i(0) \leq 0$, where the trajectory will stop at $\mzti=0$ (see \cref{app:sec:bridged_dynamics}). We observe that if 
\begin{align*}
    \lambdayxi \lambda_i^{-1} - \tilde{\epsilon}_i \lambda_i^{-1/2} \geq \modelz_i^* \geq \modelz_i(0),  
\end{align*}
or
\begin{align*}
     \modelz_i(0) > \modelz_i^* \geq \lambdayxi \lambda_i^{-1} - \tilde{\epsilon}_i \lambda_i^{-1/2},  
\end{align*}
then
\begin{align*}
        \frac{d}{dt}\lossgeni(t) \leq  0, \; \forall t,
\end{align*}
and so the error curve is monotonically decreasing in $t$. If instead
\begin{align*}
    \modelz_i^* \geq \modelz_i(0) \geq \lambdayxi \lambda_i^{-1} - \tilde{\epsilon}_i \lambda_i^{-1/2},  
\end{align*}
or
\begin{align*}
    \lambdayxi \lambda_i^{-1} - \tilde{\epsilon}_i \lambda_i^{-1/2} \geq \modelz_i(0) > \modelz_i^*,  
\end{align*}
then,
\begin{align*}
        \frac{d}{dt}\lossgeni(t) \geq  0, \; \forall t,
\end{align*}
and the error curve is monotonically increasing in $t$. Finally, if the point $\lambdayxi \lambda_i^{-1} - \tilde{\epsilon}_i \lambda_i^{-1/2}$ lies on the path between $\modelz_i(0)$ and the global minimum $\modelz_i^*$ (or alternatively, the fixed point $\mzti=0$ if $\gamma_i=0, \modelz_i(t) \leq 0$), i.e. if 
\begin{align*}
    \modelz_i^*  > \lambdayxi \lambda_i^{-1} - \tilde{\epsilon}_i \lambda_i^{-1/2}  > \modelz_i(0),  
\end{align*}
or
\begin{align*}
     \modelz_i(0)  > \lambdayxi \lambda_i^{-1} - \tilde{\epsilon}_i \lambda_i^{-1/2}  > \modelz_i^*,  
\end{align*}
the error curve will be decreasing in $t$ up until the point $\mzti = \lambdayxi \lambda_i^{-1} - \tilde{\epsilon}_i \lambda_i^{-1/2}$, after which the derivative of the error curve will change signs, and the error will become increasing in $t$. The resulting error curve hence will follow a U-shape. An exception can be seen for the case $\gamma_i=0$ and $\modelz(0) \leq 0$, where the trajectory might stop before the point $\mzti = \lambdayxi \lambda_i^{-1} - \tilde{\epsilon}_i \lambda_i^{-1/2}$. In this case, the error curve will yet again be monotonically decreasing in $t$. 

To conclude, each error curve $\lossgeni(t)$ as part of the sum in \cref{eq:full_test_loss_z}, will be either monotonically decreasing, U-shaped, or monotonically increasing in $t$. Subsequently, the full generalisation curve ($\lossgen(t)$) is a sum of monotonically decreasing, U-shaped, or monotonically increasing curves. This finishes the proof of \cref{prop:test_mse}.

\subsection{Proof of \cref{lemma:convex_minima}}
\label{app:sec:convex_minima}

We provide a proof of \cref{lemma:convex_minima}. First, following the dynamics in \cref{eq:two_layer_dynamics_z_relaxed}, we note that the second derivative of the weight $\modelz(t)$ with respect to $t$ is
\begin{align*}
    \frac{d^2\modelz(t)}{dt^2} = 4\lr^2 \modelz(t) \diffz^2 - \lambda \big(\mykb \big)\diffz,
\end{align*}
dropping all subscripts $i$ for notational clarity. 

The first and second time derivatives of $ \lossgeni(t)$  follow as
\begin{align}
    \frac{d}{d t} \lossgeni(t) & = -2\lambda \diffzt \frac{d\modelz(t)}{dt} \nonumber \\ &= -2\lambda \diffzt \sqrt{\mykb} (\lambdayx-\lambda \modelz(t)),
    \label{eq:app:first_time_der}\\ 
    \frac{d^2}{d t^2} \lossgeni(t) &= 2\lambda \left(\frac{d\modelz(t)}{dt} \right)^2 - 2\lambda \diffzt \frac{d^2\modelz(t)}{dt^2} \nonumber \\ & = 2\lambda \diffz \Big(\big(\mykb\big) \big(\lambdayx + \lambda \truez - 2 \lambda \mzt \big) \nonumber \\ & \eqspace \quad - 4\lr^2\modelz(t) \diffzt \diffz \Big).
    \label{eq:app:second_time_der}
\end{align}
Although we drop all other subscripts $i$, we keep the subscript on the error function ($\lossgeni(t)$) to emphasise that we consider the error curve for a single weight $\mzt$.

Evaluated at $\modelz(t)=\truez$, the second derivative is
\begin{align}
    & \frac{d^2}{d t^2} \lossgeni(t) \Big|_{\mzt = \truez} = 
   \frac{ 2\rho^2\lambdayx^2 \big(\gamma^2 \lambda^2 + 4\lr^2 \lambdayx^2(1-\rho)^2\big)}{\lambda},
   \label{eq:app:second_time_der_minima}
\end{align}
using $\truez = (1-\rho)\lambdayx / \lambda$. We immediately see that this derivative is positive if $1 \geq \rho > 0$, with an exception for the case $\rho=1$ if $\gamma=0$, but note that this scenario is not included under \assump. Hence, the error curve is convex in $t$ under \assump and if $1 \geq \rho > 0$.
 
For $\rho=0$, to see that the curve is convex in $t$ leading up to the point $\truez$, note that solving
\begin{align*}
    \frac{d^2}{d t^2} \lossgeni(t) = 0,
\end{align*}
with $\truez=\starz$, yields two potential inflection points of the error curve, located at
\begin{align*}
    \hat{z}^{\pm} = \frac{\lr \lambdayx \pm  \sqrt{\lr^2 \lambdayx^2-6 \gamma^2 \lambda^2}}{6 \lambda \lr}. 
\end{align*}
(the other solutions are undulation points). 
The roots $\hat{z}^{-}$ and $\hat{z}^{+}$ are real if $\lr^2 \lambdayx^2 \geq 6 \gamma^2 \lambda^2$ (including the case $\gamma=0$). In this case, observe that we can upper bound the square root $\sqrt{\lr^2 \lambdayx^2-6 \gamma^2 \lambda^2} \leq \lr \lambdayx$, and so
\begin{align*}
    \hat{z}^{\pm} \leq \frac{\lambdayx}{3\lambda}.
\end{align*}
Observe that this upper bound, under \assump, is passed before the true minimum at $\truez=\starz=\lambdayx/\lambda$. Evaluating the second derivative at a point beyond the upper bound (but before the minimum $\truez$), we find
\begin{align*}
   \frac{d^2}{d t^2} \lossgeni (t) \Big|_{\mzt = \frac{\lambdayx}{2\lambda}} = \gamma^2 \lambda \lambdayx^2+\frac{\lr^2 \lambdayx^4}{2 \lambda} > 0, 
\end{align*}
with the inequality following from \assump. Hence, the curve is convex in $t$ leading up to the minimum $\truez$. 

If instead $\lr^2 \lambdayx^2 < 6 \gamma^2 \lambda^2$ (including the case $\lr=0$), then the roots $\hat{z}^{-}$ and $\hat{z}^{+}$ are complex, and so there are no inflection points in the error curve. As we have shown that the curve is convex in $t$ at one point (namely at the point $\mzt = \lambdayx/(2\lambda)$), the curve must be convex in $t$ on the full path, and therefore, again, it must be convex leading up to the true minimum $\truez$. Moreover, for $\rho=0$, the true minimum $\truez$ is equal to the global minimum $\modelz^*$, which is reached at $t = \infty$ and where the weight $\mzt$ remains once reached. Hence, although the second derivative is $0$ at $\truez$ if $\rho=0$ (see \cref{eq:app:second_time_der_minima}), the error curve will not change convexity at this point and, so, the true minimum is an undulation point on $\lossgeni(t)$. This concludes the proof of \cref{lemma:convex_minima}.

\subsection{Proof of \cref{lemma:inflection_points}}
\label{app:sec:infl_points}
For proving \cref{lemma:inflection_points}, we seek the inflection points of the error curve in \cref{eq:test_loss_single_curve}, with $\modelz(t)$ following the dynamics in \cref{eq:two_layer_dynamics_z_relaxed}. As previously, we perform our derivations under \assump, introduced in the main paper. In addition, and as in previous derivations, we drop all subscripts $i$ for notational clarity (with an exception for the subscript on the error, $\lossgeni(t)$). 

The first and second time derivatives of $\lossgeni(t)$ are given in \cref{eq:app:first_time_der,eq:app:second_time_der}. Before moving on to analysing the inflection points of the error curve, we evaluate its second derivative at the point $\mzt = 0$:
\begin{align}
    &\frac{d^2}{d t^2} \lossgeni(t) \Big|_{\mzt = 0} = 2\gamma^2 \lambda \lambdayx^2(2-\rho).
    \label{eq:app:second_time_der_0}
\end{align}
We observe that under \assump, the second derivative is positive (or zero) at this point. 

For analysing the inflection points, we solve the equation
\begin{align*}
    \frac{d^2}{d t^2} \lossgeni(t) = 0, 
\end{align*}
with respect to $\modelz(t)$. This yields
\begin{align*}
    2\diffz = 0,
\end{align*}
or
\begin{align*}
    &\lambda \Big(\big(\mykb\big) \big(\lambdayx + \lambda \truez - 2 \lambda \mzt \big) - 4\lr^2\modelz(t) \diffzt \diffz \Big)  = 0.
\end{align*}
The solution to the first equation is $\modelz(t) = \starz$. This is not an inflection point, but instead an undulation point. The reason is that $\modelz(t)$ will not grow beyond its global minimum, and hence, the curve will not change convexity at this point.
We instead continue by analysing the second equation, amounting to analysing the roots of the polynomial
\begin{align*}
    &f(\mzt) = a_z \mzt^3 + b_z \mzt^2 + c_z \mzt + d_z, \\
    & a_z = \az, \\
    & b_z = \bz, \\
    & c_z = \cz, \\
    & d_z = \dz.
\end{align*}
We observe that the polynomial $f(\mzt)$ with $\gamma \neq 0, \lr > 0$ is cubic in $\mzt$, and has a maximum of three real roots, possibly corresponding to inflection points in the error curve $\lossgeni(t)$. As there are no other possible inflection points, the error curve $\lossgeni(t)$ has a maximum of three inflection points on the interval $[0, \modelz^*]$. This finishes the proof of the first part of \cref{lemma:inflection_points}.

For analysing the number and relative positions of inflection points on $\lossgeni(t)$, we will separate our further analysis into three cases: (i) $\gamma \neq 0, \lr > 0$, (ii) $\gamma \neq 0, \lr = 0$, (iii) $\gamma = 0, \lr > 0$.

\paragraph{Case $\boldsymbol{\gamma \neq 0, \lr > 0}$.} We start with the most general case, assuming $\gamma \neq 0$ and $\lr > 0$. We will carry out an analysis of the roots to the polynomial $f(\mzt)$ with respect to the model weight $\mzt$, identifying the number of roots as well as their positions relative to the true minimum $\truez$. Note that, as $\mzt$ is continuous and growing in $t$ under assumptions (i)-(iv), the analysis is directly transferable to the relative position and number of roots of the polynomial in $t$. 

First, we observe that $f(\mzt)$ has three distinct real roots if its \textit{discriminant}
\begin{align*}
    \Delta_f &= 18 a_z b_z c_z d_z - 4 a_z c_z^{3}  -27a_z^{2} d_z^{2} + b_z^{2}c_z^{2} - 4 b_z^{3}d_z \\
    & = -16 \lambda^2 \lr^2 \Big(24 \gamma^6 \lambda^6 + \gamma^4 \lambda^4 \lr^2 \lambdayx^2 \big(11 \rho^2-188 \rho+188\big)  \\
    & \eqspace + 16 \gamma^2 \lambda^2 \lr^4 \lambdayx^4  \big(8 \rho^4-33 \rho^3+55 \rho^2-44 \rho+22\big)  \\
    & \eqspace \eqspace - 64 \lr^6 
 \lambdayx^6 (\rho-1)^2 \big(\rho^2-\rho+1\big) \Big) 
\end{align*}
is positive, i.e. if $\Delta_f > 0$. If instead $\Delta_f = 0$, then at least two of the three roots are the same. In the case $\Delta_f < 0$, the polynomial $f(\mzt)$ has only one real root. 

We will use \textit{Rouths algorithm} to show that all of the roots of the polynomial $f(\mzt)$ have strictly positive real parts. For a cubic polynomial on the form
\begin{align*}
a_0 x^3 + b_0 x^2 + a_1 x + b_1, 
\end{align*}
Rouths algorithm analyses the following table:

\begin{center}
    \begin{tabular}{c c}
        $a_0$ & $a_1$ \\
        $b_0$ & $b_1$ \\
        $c_0$ & $0$ \\
        $d_0$ & $0$
    \end{tabular}
\end{center}
with 
\begin{align*}
    &c_0 = \frac{b_0 a_1 - b_1 a_0}{b_0}, \\
    &d_0 = b_1.
\end{align*}
Based on this table, Rouths theorem states that if the coefficient $a_0$ is strictly positive, i.e. if $a_0 > 0$, then all roots of the polynomial has a strictly negative real part, if and only if the coefficients $a_0, b_0, c_0$ and $d_0$ are all strictly positive. The number of roots with strictly positive real parts is equal to the number of times that the sequence $a_0, b_0, c_0, d_0$ changes sign. 

For the polynomial $f(\mzt)$, we have
\begin{align*}
    & a_0 = a_z = \az, \\
    & b_0 = b_z = \bz, \\
    & c_0 = \frac{b_z c_z - d_z a_z}{b_z} = \frac{\gamma^2 \lambda^2}{2} + 4 \lr^2 \lambdayx^2 (1-\rho), \\
    & d_0 = d_z = \dz.
\end{align*}
We see directly that under \assump and if $\gamma \neq 0, \lr > 0$, then: $a_0 > 0$, $b_0 < 0, c_0 > 0, d_0 < 0$. Hence, the sequence  $a_0, b_0, c_0, d_0$ changes sign three times. Following Rouths theorem, the three roots of the polynomial therefore all have strictly positive real parts. Subsequently, any real root of $f(\mzt)$ lies after the point $\mzt = 0$. 

Note that, for $\gamma \neq 0, \lr > 0$, and under \assump, the error curve $\lossgeni(t)$ is convex in $t$ at $\mzt = 0$ and $\mzt = \truez$ (its second derivative is positive at these points, see \cref{eq:app:second_time_der_minima,eq:app:second_time_der_0}) if $\rho \in (0, 1]$. For $\rho \in (0, 1]$, we hence must have an even number of inflection points on the interval $(0, \truez)$. In the case where the polynomial $f(\mzt)$ has one real root, i.e. when $\Delta_f < 0$, this root, if corresponding to an inflection point, must therefore lie after the point $\truez$. In this case, we would have zero inflection points on the interval $(0, \truez)$. However, if the polynomial has three distinct real roots, i.e. if $\Delta_f > 0$, another possibility is that two, out of the three possible, inflection points lie on the interval $(0, \truez)$. 

When $\rho=0$, the error curve $\lossgeni(t)$ is not convex in $t$ at $\mzt=\truez$ (although it is still convex at $\mzt=0$), but $\truez$ is instead an undulation point (see \cref{app:sec:convex_minima}). In this case, we can show that one of the three possible roots of the polynomial $f(\mzt)$ is equal to $\starz$, and $\lossgeni(t)$ has a maximum of two inflection points, see the derivation in \cref{app:sec:convex_minima}. Since the true minimum $\truez$ and the global minimum $\starz$ coincide for $\rho=0$, any relevant inflection points must lie in the interval $(0, \truez)$. We have shown that the error curve when $\rho=0$ is convex in $t$ at $\mzt=0$ (when $\gamma \neq 0$, see \cref{eq:app:second_time_der_0}) as well as leading up to the point $\mzt=\truez$ (see \cref{app:sec:convex_minima}). Hence, as with $\rho \in (0, 1]$, we must have either zero or two inflection points on $(0, \truez)$. Considering the two cases $\rho \in (0, 1]$ and $\rho=0$ together, we conclude that for $\gamma \neq 0, \lr > 0$ and under \assump, a \textit{maximum} of two inflection points lie in the interval $(0, \truez)$.

We continue our investigation by considering the case where the polynomial $f(\mzt)$ has three real roots, i.e. the case for which $\Delta_f > 0$. For this analysis, we assume $\rho > 0$, as we have already concluded that if $\rho=0$, the error curve has a maximum of two inflection points. Our aim is to analyse the relative position of the roots of $f(\mzt)$ to the point $\truez$. To this end, we shift the function $f(\mzt)$ by $\truez$, using a change of variables, 
\begin{align*}
    &g(\mxt) = a_x \mxt^3 + b_x \mxt^2 + c_x \mxt + d_x, \\
    & a_x = \ax, \\
    & b_x = \bx, \\
    & c_x = \cx, \\
    & d_x = \dx,
\end{align*}
with $\mxt = \mzt - \truez$, such that $g(0)=f(\truez)$. Note that $g(\mxt)$ has equally many distinct roots in $\mxt$ as $f(\mzt)$ has in $\mzt$, as shifting the function along the x-axis does not change this fact. Moreover, when $\Delta_f > 0$, for which $f(\mzt)$ has three real and distinct roots, since all roots of $f(\mzt)$ are strictly positive, the roots of $g(\mxt)$ must be larger than $-\truez$.

We evaluate $g(\mxt)$ at the points $\mxt = - \truez$ (corresponding to $\mzt = 0$) and $\mxt = 0$ (corresponding to $\mzt = \truez$)
\begin{align*}
    &g(-\truez) = \dz, \\
    &g(0) = \dx. 
\end{align*}
Hence, under \assump and with $\gamma \neq 0, \lr > 0$, $\rho \in (0, 1]$, we have $g(-\truez), g(0) < 0$. Now, if we can find a point on the interval $(-\truez, 0)$ such that $g(\mxt)>0$, then we must conclude that there are at least two roots of $g(\mxt)$ lying on the interval $(-\truez, 0)$ (as the function in that case must change sign two times in between $0$ and $\truez$), and therefore two roots of $f(\mzt)$, lying on the interval $(0, \truez)$. 

For this purpose, observe that if $g(\mxt)$ has three distinct real roots (i.e. if $\Delta_f  > 0$), then the polynomial must also have two distinct and real local optima; one local minimum and one local maximum, lying in between the roots of the polynomial. The local optima are
\begin{align*}
    x^{\pm} = \frac{2 \lr \lambdayx (7 \rho - 5) \pm \sqrt{4 \lr^2 \lambdayx^2 \big(4 \rho^2-7 \rho+7\big) - 18 \gamma^2 \lambda^2}}{18 \lr \lambda}.
\end{align*}
For $\gamma \neq 0, \lr > 0$, we have $a_x > 0$, and, therefore $g(\mxt) \rightarrow - \infty$ if $\mxt \rightarrow - \infty$ and $g(\mxt) \rightarrow \infty$ if $\mxt \rightarrow \infty$. Hence, the local maximum must be located at a point before (in time) the local minimum, i.e. the local maximum is the point
\begin{align*}
     x^{-} = \frac{2 \lr \lambdayx (7 \rho - 5) - \sqrt{4 \lr^2 \lambdayx^2 \big(4 \rho^2-7 \rho+7\big) - 18 \gamma^2 \lambda^2}}{18 \lr \lambda}.
\end{align*}
Clearly, if $5/7 > \rho > 0$, then  $x^- < 0$. Note also that in the case of three real and distinct roots, the polynomial $g(\mxt)$ must be positive at the local maximum, i.e. $g(x^-) > 0$. Moreover, since all of the roots of $g(\mxt)$ are larger than $-\truez$ and as $g(-\truez) < 0$, we therefore also must have $x^- > -\truez$. Hence, we have found a point at which $g(\mxt) > 0$ and that lie on the interval $(-\truez, 0)$. 

Coming back to the original polynomial, $f(\mzt)$, we conclude that if $f(\mzt)$ has three real and distinct roots and if $5/7 > \rho > 0$, then at least two of them must lie in the interval $(0, \truez)$. If all of the three roots of $f(\mzt)$ correspond to inflection points in the curve $\lossgeni(t)$, and as we have concluded that there can be a maximum of two roots on the interval $(0, \truez)$, then exactly two of these must lie in the interval $(0, \truez)$, and the third on the interval $(\truez, \starz)$.

\paragraph{Case $\boldsymbol{\gamma \neq 0, \lr=0}$.}
For the one-layer model, with $\lr=0$, the polynomial $f(\mzt)$ simplifies to
\begin{align*}
    &f(\mzt) = 2\gamma^2 \lambda^2 \mzt \dz.
\end{align*}
Solving $f(\mzt) = 0$, we find
\begin{align*}
    \hat{\modelz} = \frac{\lambdayx(2-\rho)}{2\lambda},
\end{align*}
corresponding to a single, potential inflection point. Note that for $\rho = 0$, we have $\hat{\modelz} = \starz$, and, as $\mzt$ never grows beyond its global minimum $\starz$, the error curve $\lossgeni(t)$ has no inflection points.

For $\rho \in (0, 1]$, it is easily verified that
\begin{align*}
    1 > \frac{2-\rho}{2} > 1-\rho,
\end{align*}
and therefore $\hat{\modelz} \in (\truez, \starz)$. Moreover, under \assump, and when $\lr=0$, $\gamma , \rho > 0$, we have
\begin{align*}
    \frac{d^2}{d t^2} \lossgeni(t) \Big|_{\mzt = \truez} = 
    2\gamma^2  \lambda \lambdayx^2 \rho^2 > 0.
\end{align*}
To show that we have a change of convexity at $\hat{\modelz}$, we take a point lying on the interval $(\hat{\modelz}, \starz)$ and evaluate the second derivative of the error at that point:
\begin{align*}
    \frac{d^2}{d t^2} \lossgeni(t) \Big|_{\mzt = \frac{\lambdayx(2-0.5\rho)}{2\lambda}} = 
    - \frac{\gamma^2  \lambda \lambdayx^2 \rho^2}{4} < 0.
\end{align*}
Hence, under \assump, the error curve $\lossgeni(t)$ is concave beyond the root $\hat{\modelz}$, indicating that the point $\hat{\modelz} \in (\truez, \starz)$ is indeed an inflection point of the curve. In terms of $t$, and initialising at $\modelz(0)$, the inflection point $\hat{\modelz}$ corresponds to
\begin{align*}
    t(\hat{z}) = \frac{\log\left(\frac{2(\lambdayx - \lambda \modelz(0))}{\lambdayx \rho} \right)}{|\gamma| \lambda},
\end{align*}
using \cref{eq:time_to_min} but replacing $\truez$ with $\hat{z}$. This inflection point lies in the interval $(t^{(1-\rho)\starz}, \infty)$. 

We conclude that under \assump and when $\lr=0$, the error curve has either zero inflection points (if $\rho=0$) or one inflection point (if $\rho \in (0,1]$), lying on the interval $(\truez, \starz)$.

\paragraph{Case $\boldsymbol{\gamma=0, \lr> 0}$.}
For the balanced two-layer dynamics ($\gamma=0$), note that the case $\rho=1$ is not covered by \assump and hence we carry out the following analysis for $\rho \in [0, 1)$. With $\gamma=0$, the polynomial $f(\mzt)$ simplifies to
\begin{align*}
    &f(\mzt) = 4 \lr^2 \mzt \big(3\lambda^2 \mzt^2 - 2\lambda \lambdayx (2-\rho) \mzt +  \lambdayx^2(1-\rho) \big).
\end{align*}
First, we note that the solution $\hat{\modelz}=0$ of $f(\mzt)=0$ is not an inflection point, as $\mzt \geq 0$. To find potential inflection points, we instead solve 
\begin{align*}
    3\lambda^2 \mzt^2 - 2\lambda \lambdayx (2-\rho) \mzt +  \lambdayx^2(1-\rho) = 0,
\end{align*}
with solutions
\begin{align*}
      \hat{\modelz}^{\pm} = \frac{\lambdayx \big(2-\rho \pm \sqrt{\rho^2- \rho +1}\big)}{3 \lambda}. 
\end{align*}
For the first root, $\hat{\modelz}^{-}$, and with $\rho \in [0, 1)$, we can verify
\begin{align*}
    1- \rho > \frac{2-\rho - \sqrt{\rho^2- \rho +1}}{3} > 0. 
\end{align*}
Therefore, $\hat{\modelz}^- \in (0, \truez)$ when $\rho \in [0, 1)$. For the second root, $\hat{\modelz}^{+}$, observe that $\hat{\modelz}^{+} = \starz$ when $\rho=0$, and hence, by similar reason as previously, $\hat{\modelz}^{+}$ is not an inflection point if $\rho=0$. For $\rho \in (0, 1)$, we find 
\begin{align*}
     1 > \frac{2-\rho + \sqrt{\rho^2- \rho +1}}{3} > 1- \rho, 
\end{align*}
and, hence, $\hat{\modelz}^+ \in (\truez, \starz)$.

We start by verifying the convexity of $\lossgeni(t)$ in between the root  $\hat{\modelz}^{-}$ and $\hat{\modelz}^{+}$. To this end, we take a point on the interval $(\hat{\modelz}^{-}, \hat{\modelz}^{+})$ and evaluate the second derivative at this point:
\begin{align*}
    \frac{d^2}{d t^2} \lossgeni(t) \Big|_{\mzt = \frac{\lambdayx(2-\rho)}{3\lambda}} = -\frac{8\lr^2 \lambdayx^4 (\rho-2) (\rho+1) (\rho^2-\rho+1) }{27 \lambda} > 0. 
\end{align*}
In other words, $\lossgeni(t)$ is convex in between the roots.

For our further analysis, we start with the case $\rho \in (0, 1)$, where both of the roots $\hat{\modelz}^{-}$ and $\hat{\modelz}^{+}$ are potential inflection points. In this scenario, we observe that the error curve is concave outside of the interval $(\hat{\modelz}^{-}, \hat{\modelz}^{+})$. We use the upper bound $1 > \sqrt{1-\rho+\rho^2}$, to find a point lying on the interval $(0, \hat{\modelz}^{-})$, and for which the second derivative of $\lossgeni(t)$ is negative:
\begin{align*}
    \frac{d^2}{d t^2} \lossgeni(t) \Big|_{\mzt = \frac{\lambdayx(1-\rho)}{3\lambda}} = -\frac{8 \lr^2 \lambdayx^4 \rho (1-\rho)^2 (\rho+2) }{27 \lambda} < 0. 
\end{align*}
Similarly, we find a point lying on the interval $(\hat{\modelz}^{+}, \starz)$, and for which the second derivative is negative:
\begin{align*}
     \frac{d^2}{d t^2} \lossgeni(t) \Big|_{\mzt = \frac{\lambdayx(3-\rho)}{3\lambda}} = -\frac{8 \lr^2 \lambdayx^4 \rho^2 (\rho-1) (\rho-3) }{27 \lambda} < 0. 
\end{align*}
Hence, for the balanced two-layer dynamics, under \assump and when $\rho \in (0, 1)$, the error curve $\lossgeni(t)$ has two inflection points: $\hat{\modelz}^{-} \in (0, \truez)$ and $\hat{\modelz}^{+} \in (\truez, \infty)$.

When $\rho=0$, we have only one root of $f(\mzt)$ that is a potential inflection point. This root is
\begin{align*}
    \hat{\modelz}^{-} = \frac{\lambdayx}{3\lambda}.
\end{align*}
We have shown already that, including when $\rho=0$, the error curve $\lossgeni(t)$ is convex at a point in the interval $(\hat{\modelz}^{-} , \starz)$ (as for $\rho=0$ we have $\hat{z}^+=\starz$, see the evaluation of the second derivative at $z(t) = \lambdayx(2-\rho)/(3\lambda)$). To show that $\hat{\modelz}^{-}$ is an inflection point when $\rho=0$, we additionally evaluate the second derivative of $\lossgeni(t))$ at a point in the interval $(0, \hat{\modelz}^{-})$: 
\begin{align*}
    \frac{d^2}{d t^2} \lossgeni(t) \Big|_{\mzt = \frac{\lambdayx}{4\lambda}} =  -\frac{9\lr^2 \lambdayx^4}{32\lambda} < 0. 
\end{align*}
Hence, for $\rho=0$, the error curve $\lossgeni(t)$ has one inflection point at $\hat{\modelz}^{-} \in (0, \truez)$.

In terms of $t$, initialising at $\modelz(0)$ the inflection points $\hat{\modelz}^{\pm}$, using \cref{eq:time_to_min} (but replacing $\truez$ with either of the roots), translates to
\begin{align*}
    t(\hat{z}^{\pm}) = \frac{\log \left(\displaystyle\frac{\big(\lambdayx -\lambda \modelz(0)\big)\big(1\pm\sqrt{\rho^2-\rho+1}\big) }{\lambda\rho \modelz(0)}\right)}{2 \lr \lambdayx},
\end{align*}
if $\rho \in (0, 1)$. If $\rho=0$, we instead have the inflection point
\begin{align*}
    t(\hat{z}^{-}) = \frac{\log \left(\displaystyle\frac{\lambdayx - \lambda \modelz(0)}{2\lambda \modelz(0)}\right)}{2 \lr \lambdayx}.
\end{align*}
If $\modelz(0) < \hat{z}^{-}$ and $\rho \in [0, 1)$, the point $t(\hat{z}^{-})$ will be an inflection point lying on the interval $(0, t^{(1-\rho)\starz})$. However, when $\modelz(0) \geq \hat{z}^{-}$, the point $t(\hat{z}^{-})$ will either be equal to $0$ (if $\modelz(0) = \hat{z}^{-}$), corresponding to an undulation point and not an inflection point, or lie outside of the interval $(0, t^{(1-\rho)\starz})$ (when $\modelz(0) > \hat{z}^{-}$), such that an inflection point corresponding to $\hat{z}^-$ does not exist. Meanwhile, when $\rho \in (0, 1)$, $t(\hat{z}^{+})$ will be an inflection point lying on the interval $(t^{(1-\rho)\starz}, \infty)$.\\

In summary, we conclude that, under \assump, the error curve $\lossgeni(t)$ has a maximum of three inflection points, whereof a maximum of two lies in the interval $(0, \truez)$. If three inflection points exist and $5/7 > \rho > 0$, then two inflection points lie in the interval $(0, \truez)$ and one in the interval $(\truez, \modelz^*)$. Under \assump, we note that $\mzt$ is continuous as well as growing in time, $t$, and hence the conclusions are directly transferable to the corresponding inflection points in terms of $t$, replacing $\truez$ with $t(\truez) = t^{(1-\rho)\starz}$ and $\starz$ with $t(\starz) = \infty$. Indeed, we can have $\modelz(0) > 0$, and it is therefore possible that there is a fewer number of inflection points on the time interval $(0, \infty)$ compared to the variable interval $(0, \starz)$. However, note that the conclusions still hold, referring to a maximum number of inflection points, or being contingent on having exactly three inflection points. With this, we end the proof of \cref{lemma:inflection_points}.
\subsection{Proof of \cref{prop:double_descent_necessary_general,prop:double_descent_necessary}}
\label{app:sec:dd_necessary}

To prove \cref{prop:double_descent_necessary_general,prop:double_descent_necessary}, we consider the weight matrix $\Modelz(t)$ with $|\activeset|$ active weights, following the dynamics in \cref{eq:two_layer_dynamics_z_relaxed}. The total generalisation error, as given by \cref{eq:full_test_loss_z}, is a sum over $|\activeset|$ error curves. We show that, under \assump, a necessary condition for this generalisation error to exhibit a double descent pattern over the course of learning, is that we can find at least one inflection point, $\hat{t}$, belonging to either one of the $|\activeset|$ individual error curves, such that 
\begin{align*}
    \min \{t_i^{(1-\rho_i)\modelz_i^*}; i \in \activeset \} < \hat{t} < \max \{t_i^{(1-\rho_i)\modelz_i^*}; i \in \activeset \},
\end{align*}
with $t_i^{(1-\rho_i)\modelz_i^*}$ defined in \cref{eq:time_to_min}. We prove this necessary condition by showing that if there is no inflection point $\hat{t}$ lying in the interval $\big({\min} \{t_i^{(1-\rho_i)\modelz_i^*}; i \in \activeset \}, \, \max \{t_i^{(1-\rho_i)\modelz_i^*}; i \in \activeset \}\big)$, then the error curve $\lossgen(t)$ does \textit{not} exhibit a double descent pattern.

We first observe that double descent requires that the error curve $\lossgen(t)$ declines in $t$, i.e. that 
\begin{align*}
    \frac{d}{dt}\lossgen(t) < 0,
\end{align*}
at a time interval succeeding an interval for which $\lossgen(t)$ grows in $t$, i.e. for which
\begin{align*}
    \frac{d}{dt}\lossgen(t) > 0.
\end{align*}
We show that this is not possible if there is no inflection point $\hat{t}$ lying in the interval \\ $\big({\min} \{t_i^{(1-\rho_i)\modelz_i^*}; i \in \activeset \}, \, \max \{t_i^{(1-\rho_i)\modelz_i^*}; i \in \activeset \}\big)$.

First, observe that in the time interval $\big[0, \min \{t_i^{(1-\rho_i)\modelz_i^*}; i \in \activeset \}\big]$, all individual error curves $\lossgeni(t)$, for $i \in \activeset$, are decreasing or constant in $t$, i.e.
\begin{align*}
    \frac{d}{dt} \lossgeni(t) \leq 0, \forall i \in \activeset,
\end{align*}
(see \cref{app:sec:test_mse}). Following \cref{eq:full_test_loss_z}, we therefore have
\begin{align*}
    \frac{d}{dt}\lossgen(t) \leq 0,
\end{align*}
on this interval. In addition, in the interval $\big[\max \{t_i^{(1-\rho_i)\modelz_i^*}; i \in \activeset \}, \infty\big)$, we have
\begin{align*}
    \frac{d}{dt} \lossgeni(t) \geq 0, \forall i \in \activeset,
\end{align*}
and therefore 
\begin{align*}
    \frac{d}{dt}\lossgen(t) \geq 0.
\end{align*}
To determine the behaviour of the generalisation error in between the two points $\min \{t_i^{(1-\rho_i)\modelz_i^*}; i \in \activeset \}$ and $\max \{t_i^{(1-\rho_i)\modelz_i^*}; i \in \activeset \}$, we observe that under \assump and following \cref{lemma:convex_minima}, each individual error curve $\lossgeni(t)$, for $i \in \activeset$ and with corresponding $\rho_i \in (0, 1]$, is convex at $\truez$. We point out that for $\rho_i=1$, we have $\truez_i=0$, and hence, under \assump, $\mzti$ must begin at the point $\truez$, so the error curve will be convex to start. For $\rho_i=0$, again following \cref{lemma:convex_minima}, the curve is convex leading up to the minimum, at which point the second derivative of $\lossgeni(t)$ is $0$, and after which $\mzti$ will remain constant (note, however, that this happens at $t = \infty$).

Taken together, if there is no inflection point $\hat{t}$, belonging to either one of the error curves $\lossgeni(t)$, for $i \in \activeset$, and lying in the interval $\big({\min} \{t_i^{(1-\rho_i)\modelz_i^*}; i \in \activeset \}, \, \max \{t_i^{(1-\rho_i)\modelz_i^*}; i \in \activeset \}\big)$, then, on this interval, we must have
\begin{align*}
    \frac{d^2}{dt^2}\lossgeni(t) \geq 0, \forall i \in \activeset,
\end{align*}
and, therefore, 
\begin{align*}
    \frac{d^2}{dt^2}\lossgen(t) \geq 0.
\end{align*}
Now, assume that the interval $\big({\min} \{t_i^{(1-\rho_i)\modelz_i^*}; i \in \activeset \}, \, \max \{t_i^{(1-\rho_i)\modelz_i^*}; i \in \activeset \}\big)$ has no inflection points (belonging to either of the individual error curves), and take any time point $t'$ for which 
\begin{align*}
    \frac{d}{dt}\lossgen(t) \Big|_{t=t'} > 0,
\end{align*}
i.e. the error curve $\lossgen(t)$ is growing at time $t'$. By the previous statements made regarding the first and second time derivatives of $\lossgen(t)$, we must have $t' \in \big({\min} \{t_i^{(1-\rho_i)\modelz_i^*}; i \in \activeset \}, \infty \big)$. Then, take any other time point $t''$ such that $t'' > t'$. Also, $t''$ must lie on the interval $ \big({\min} \{t_i^{(1-\rho_i)\modelz_i^*}; i \in \activeset \}, \infty \big)$. If $t'' \in \big({\min} \{t_i^{(1-\rho_i)\modelz_i^*}; i \in \activeset \}, \, \max \{t_i^{(1-\rho_i)\modelz_i^*}; i \in \activeset \}\big)$, it must hold, as we have concluded that $\lossgen(t)$ is convex on this interval when it does not contain inflection points, that
\begin{align*}
    \frac{d}{dt}\lossgen(t) \Big|_{t=t''} \geq \frac{d}{dt}\lossgen(t) \Big|_{t=t'} > 0.
\end{align*}
If instead $t'' \in \big[{\max} \{t_i^{(1-\rho_i)\modelz_i^*}; i \in \activeset \}, \infty\big)$, then, by previous conclusions,
\begin{align*}
    \frac{d}{dt}\lossgen(t) \Big|_{t=t''} \geq 0.
\end{align*}
We can conclude that there is no time point $t''$, following a time point $t'$ at which $\lossgen(t)$ is growing in $t$, such that 
\begin{align*}
    \frac{d}{dt}\lossgen(t) \Big|_{t=t''} < 0.
\end{align*}
Therefore, the error curve $\lossgen(t)$ will not exhibit a double descent pattern, if there is no inflection point $\hat{t}$ on $\big({\min} \{t_i^{(1-\rho_i)\modelz_i^*}; i \in \activeset \}, \, \max \{t_i^{(1-\rho_i)\modelz_i^*}; i \in \activeset \}\big)$

On the other hand, assume that there exists an inflection point $\hat{t}$ (belonging to either of the individual error curves $\lossgeni(t)$) that lies in $\big({\min} \{t_i^{(1-\rho_i)\modelz_i^*}; i \in \activeset \}, \, \max \{t_i^{(1-\rho_i)\modelz_i^*}; i \in \activeset \}\big)$. Then, there will be at least one sub-interval of this interval, where it holds for at least one $\mzti$, $i \in \activeset$, that
\begin{align*}
    \frac{d^2}{dt^2}\lossgeni(t) < 0.
\end{align*}
Hence, the error curve $\lossgen(t)$ possibly exhibits a non-convex behaviour on the interval $ \big({\min} \{t_i^{(1-\rho_i)\modelz_i^*}; i \in \activeset \}, \, \max \{t_i^{(1-\rho_i)\modelz_i^*}; i \in \activeset \} \big)$, potentially giving rise to a double descent pattern. This concludes the proof of \cref{prop:double_descent_necessary_general}.

For the proof of \cref{prop:double_descent_necessary}, consider the special case with two active weights, $\modelz_i(t)$ and $\modelz_j(t)$. From the assumption  $t_i^{(1-\rho_i)\modelz_i^*} < t_j^{(1-\rho_j)\modelz_j^*}$, it follows directly that
\begin{align*}
    &\min \{t_i^{(1-\rho_i)\modelz_i^*}; i \in \activeset \} = t_i^{(1-\rho_i)\modelz_i^*},\\
    &\max \{t_i^{(1-\rho_i)\modelz_i^*}; i \in \activeset \} = t_j^{(1-\rho_j)\modelz_j^*}.  
\end{align*}
Hence, the necessary condition for observing epoch-wise double descent simplifies to having at least one inflection point $\hat{t}$, belonging to either of the error curves $\lossgeni(t)$ and $\lossgenj(t)$, with 
\begin{align*}
    t_i^{(1-\rho_i)\modelz_i^*} < \hat{t} <t_j^{(1-\rho_j)\modelz_j^*}.
\end{align*}
Now, provided that it exists, let $\hat{t}_j^{-}$ denote the maximum inflection point belonging to the second curve, $\lossgenj(t)$, lying in the interval $(0, t_j^{(1-\rho_j)\modelz_j^*})$. Moreover, let, provided that it exists, $\hat{t}_i^{+}$ denote the minimum inflection point belonging to the first error curve, $\lossgeni(t)$, and lying on the interval $(t_i^{(1-\rho_i)\modelz_i^*}, \infty)$. We note that we only need to consider the maximum inflection point on the interval $(0, t_j^{(1-\rho_j)\modelz_j^*})$ belonging to the curve $\lossgenj(t)$, as if a smaller inflection point exists on the same interval and this inflection point also lies in the interval  $(t_i^{(1-\rho_i)\modelz_i^*}, t_j^{(1-\rho_j)\modelz_j^*})$, so must $\hat{t}_j^{-}$. In a similar manner, we only need to consider the minimum inflection point on the interval $(t_i^{(1-\rho_i)\modelz_i^*}, \infty)$, belonging to the curve $\lossgeni(t)$, as if a larger inflection point exists on this interval and also lies in the interval $(t_i^{(1-\rho_i)\modelz_i^*}, t_j^{(1-\rho_j)\modelz_j^*})$, so must $\hat{t}_i^{+}$. Then, a necessary condition for double descent, following the condition above, is that at least one of the two inflection points $\hat{t}_i^{+}$ and $\hat{t}_j^{-}$ lies in the interval  $(t_i^{(1-\rho_i)\modelz_i^*}, t_j^{(1-\rho_j)\modelz_j^*})$, which can be equivalently written as fulfilling one of
\begin{align*}
    t_i^{(1-\rho_i)\modelz_i^*} < \hat{t}_j^{-}, \\
    \hat{t}_i^{+} < t_i^{(1-\rho_i)\modelz_i^*}.
\end{align*}
Note that assuming $t_i^{(1-\rho_i)\modelz_i^*} < t_j^{(1-\rho_j)\modelz_j^*}$ is without loss of generality. If the opposite holds, just change the order of $\modelz_i(t)$ and $\modelz_j(t)$. This concludes the proof of \cref{prop:double_descent_necessary}.

\subsection{Deep linear models}
\label{app:sec:deeper_models}

We derive the approximate dynamics of decoupled multi-layer linear neural networks, given in \cref{eq:deeper_models_dynamics_z}, and identify inflection points in the error curve $\lossgeni(t)$ (\cref{eq:test_loss_single_curve}) when $\modelz_i(t)$ follows these dynamics.

\subsubsection{Derivation of dynamics}
We start by deriving the approximate dynamics in \cref{eq:deeper_models_dynamics_z}. The linear neural network of $L$ layers is defined according to
\begin{align*}
    &\hat{y} = x \Modelw^\top, \\
    &\Modelw = \prod_{\ell=1}^{L} \Modelw^{(\ell)},
\end{align*}
with weight matrices $\Modelw^{(\ell)} \in \mathbb{R}^{h^{(\ell)} \times h^{(\ell-1)}}$, for $\ell=1, 2, \ldots, L$, and where $h^{(0)} = \xdim$ and $h^{(L)} = \ydim$. 
The gradient of the MSE loss in \cref{eq:full_loss} with respect to the $\ell^{\tht}$ weight is
\begin{align*}
    \frac{1}{\lr_{\ell}}\frac{d}{dt} \Modelw^{(\ell)} = \frac{1}{n}\left(\prod_{j=\ell+1}^{L} \Modelw^{(j)}\right)^\top (\yvecdd^\top \xvecdd - \Modelw \xvecdd^\top \xvecdd) \left(\prod_{j=1}^{\ell-1} \Modelw^{(j)}\right)^\top,
\end{align*}
where $\lr_{\ell} \geq 0$ is a layer-dependent learning rate. We assume $\prod_{j=\ell_l}^{\ell_u} \Modelw^{(j)} = \mathbb{I}$ (the identity matrix) if $\ell_l > \ell_u$.

We follow \textcite{Saxe2014} and initialise weight matrices according to $\Modelw^{(\ell)}(0) = {R^{(\ell+1)}} D^{(\ell)} {R^{(\ell)}}^\top$, where $R^{(\ell)}$, for $\ell=1, 2, \ldots, L$, are orthogonal matrices and $D^{(\ell)}$, for $\ell=1, 2, \ldots, L$, are diagonal matrices. We set $R^{(1)}=\V$ and $R^{(L)} = \Uyx$. Subsequently, we define the \textit{synaptic} weights according to 
\begin{align*}
    \Modelz^{(\ell)} = {R^{(\ell+1)}}^{\top} \Modelw^{(\ell)} R^{(\ell)},
\end{align*}
for $\ell=1, 2, \ldots, L$. The $\ell^{\tht}$ synaptic weight, using the singular value decompositions $n^{-1}\xvecdd^\top \xvecdd = \V\Lambda \V^\top$ and $n^{-1}\yvecdd^\top \xvecdd = \Uyx \Syx \V^\top$ (as previously), evolves as
\begin{align*}
    \frac{1}{\lr_\ell}\frac{d}{dt} \Modelz^{(\ell)} = \left(\prod_{j=\ell+1}^{L} \Modelz^{(j)}\right)^\top \big(\Syx - \Modelz \Lambda \big) \left(\prod_{j=1}^{\ell-1} \Modelz^{(j)}\right)^\top.
\end{align*}
Observe that, with the given initialisation, synaptic weight matrices will be initialised as diagonal matrices, with $\Modelz^{(\ell)}(0) = D^{(\ell)}$, for $\ell=1, 2, \ldots, L$. With this initialisation, the gradients of any non-diagonal elements will be zero and, subsequently, each synaptic weight matrix $\Modelz^{(\ell)}(t)$ will remain diagonal throughout the course of learning. Following the above dynamics, the gradient of the $i^{\tht}$ diagonal element of layer $\ell$, denoted $a_i^{(\ell)}$, is
\begin{align*}
    \frac{1}{\lr_{\ell}}\frac{da_i^{(\ell)}}{dt} = \prod_{j \neq \ell} a_i^{(j)} \bigg(\lambdayxi - \lambda_i \prod_{j=1}^{L} a_i^{(j)} \bigg),
\end{align*}
with $\lambda_i$ and $\lambdayxi$ the $i^{\tht}$ diagonal elements of $\Lambda$ and $\Syx$, respectively.

Similar to \textcite{Saxe2014} and the two layer dynamics in \cref{eq:two_layer_dynamics_a_b_relaxed}, these dynamics exhibit conserved quantities in the form of $\lr_j {a_i^{(\ell)}}^2 - \lr_{\ell} {a_i^{(j)}}^2 = \gamma_i^{(\ell, j)}$ $\forall \ell, j \in \{1, \ldots, L\}$. This, we can easily verify as
\begin{align*}
    \frac{d}{dt} (\lr_j {a_i^{(\ell)}}^2 - \lr_{\ell} {a_i^{(j)}}^2) = 2 \lr_j a_i^{(\ell)}\frac{d a_i^{(\ell)}}{dt} - 2 \lr_{\ell} a_i^{(j)}\frac{d a_i^{(j)}}{dt} = 0.
\end{align*}
We use these conserved quantities to rewrite the gradient dynamics for the $i^{\tht}$ diagonal element, $\modelz_i = \prod_{\ell=1}^{L} a_i^{(\ell)}$,  of the full weight matrix $\Modelz =\prod_{\ell=1}^{L} Z^{(\ell)}$, according to
\begin{align*}
   \frac{d\modelz_i}{dt} &= \sum_{\ell=1}^{L} \lr_{\ell} \prod_{j \neq \ell} {a_i^{(j)}}^2 (\lambdayxi - \lambda_i z_i) \\ &= \sum_{\ell=1}^{L} \lr_\ell \prod_{j \neq \ell} \frac{\gamma_i^{(j, L)} + \lr_j {a_i^{(L)}}^2}{\lr_L}  (\lambdayxi - \lambda_i z_i).
\end{align*}
To simplify the dynamics further, we assume $L$ to be even, and group parameters into two groups, with $\gamma_i^{(\ell, L)} = \gamma_i$, $\lr_\ell = \lr_1$, for $\ell = 1, \ldots, L/2$, and with $\gamma_i^{(\ell, L)} = 0$, $\lr_\ell = \lr_L$, for $\ell = L/2 + 1, \ldots, L$. Then, 
\begin{align*}
   \frac{d\modelz_i}{dt} &= \frac{L}{2} \bigg( \lr_1 \bigg(\frac{\gamma_i + \lr_1 {a_i^{(L)}}^2}{\lr_L} \bigg)^{L/2-1} \Big( {a_i^{(L)}}^2\Big)^{L/2}   \\ & \eqspace \eqspace \eqspace + \lr_L \bigg(\frac{\gamma_i + \lr_1 {a_i^{(L)}}^2}{\lr_L}\bigg)^{L/2} \Big( {a_i^{(L)}}^2\Big)^{L/2-1} \bigg)(\lambdayxi - \lambda_i z_i).
\end{align*}
Moreover, we have
\begin{align*}
    \modelz_i = \prod_{\ell=1}^{L} a_i^{(\ell)} = \pm \left(\sqrt{\frac{\gamma_i + \lr_1 {a_i^{(L)}}^2}{\lr_L}}\right)^{L/2}\left( \sqrt{{a_i^{(L)}}^2}\right)^{L/2},
\end{align*}
from which we find 
\begin{align*}
    {a_i^{(L)}}^2 = \frac{-\gamma_i + \sqrt{\gamma_i^2 + 4 \lr_1 \lr_L \modelz_i^{4/L} }}{2\lr_1}.
\end{align*}
Inserting this into the dynamics of $\modelz_i$ and simplifying, we obtain
\begin{align*}
    \frac{dz_i}{dt} = \frac{L}{2} \sqrt{\gamma_i^2 + 4 \lr_1 \lr_L z_i^{4/L}} z_i^{2(1 - 2/L)} (\lambdayxi - \lambda_i z).
\end{align*}
For large $L$, we have
\begin{align*}
     \frac{dz_i}{dt} \approx \frac{L}{2} \sqrt{\gamma_i^2 + 4\lr^2} z_i^2 (\lambdayxi - \lambda_i z_i),
\end{align*}
where we define $\lr \coloneqq \sqrt{\lr_1 \lr_L}$. We have recovered \cref{eq:deeper_models_dynamics_z}, up to renaming the learning rates $\lr_1$ and $\lr_L$ to $\lr_a$ and $\lr_b$, respectively.

\subsubsection{Generalisation error}
We study the generalisation error of the multi-layer linear model, using the approximate dynamics in \cref{eq:deeper_models_dynamics_z}. Just as in the case of the two-layer linear model, we assume that the true model is linear, as given by \cref{eq:true_model}. Then, we note that we can rewrite the generalisation error of the multi-layer linear model as a sum over individual error curves, just as for the two-layer linear model, as in \cref{eq:full_test_loss_z}. 

To understand the behaviour of the generalisation error of the weight matrix $\Modelz(t)$ with weights following the decoupled dynamics of \cref{eq:deeper_models_dynamics_z}, we analyse the behaviour of the error curve in \cref{eq:test_loss_single_curve}, under \assump. Adapted to the multi-layer dynamics,  we require, for weight $\mzti$ to be active, that the corresponding \textit{effective} learning rate $\sqrt{\gamma_i^2 + 4\lr^2}$ is non-zero as well as that $\modelz_i(0) > 0$. 

We analyse the shape of the error curve given in $\cref{eq:test_loss_single_curve}$, by identifying its inflection points. From here on, we will, as previously, drop all subscripts $i$ (with an exception for $\lossgeni(t)$). First, for simplification, and as the effective learning rate is constant, we define $\bar{\lr} \coloneqq \sqrt{\gamma_i^2 + 4\lr^2}$. Then, from \cref{eq:deeper_models_dynamics_z}, we have
\begin{align*}
    \frac{d^2\modelz}{dt^2} = \frac{1}{4}L^2 \bar{\lr}^2 z^3 \big(\lambdayx - \lambda \modelz)(2\lambdayx - 3\lambda \modelz).
\end{align*}
The first and second time derivatives of the error curve in \cref{eq:test_loss_single_curve}, with $\mzt$ following the dynamics in \cref{eq:deeper_models_dynamics_z}, are
\begin{align*}
    &\frac{d\lossgeni(t)}{dt} = - L \bar{\lr} \modelz (t)^2 \big((1-\rho)\lambdayx - \lambda \modelz (t)\big)\big(\lambdayx - \lambda \modelz (t) \big), \\
    &\frac{d^2\lossgeni(t)}{dt^2} = -\frac{1}{2}L^2 \bar{\lr}^2 \mzt^3 \big(\lambdayx - \lambda \mzt  \big)\big(4\lambda^2 \mzt^2 - 3\lambda \lambdayx (2-\rho) \mzt + 2\lambdayx^2(1-\rho)\big).
\end{align*}
We point out that under \assump and with $\bar{\lr},\, \modelz(0) > 0$, it holds that
\begin{align*}
    \lambdayx - \lambda \modelz (t)  \geq 0, \; \forall t,
\end{align*}
and therefore, $\mzt$ is growing in $t$, i.e.
\begin{align*}
    \frac{dz}{dt} \geq 0, \; \forall t.
\end{align*}
Following this, the error curve in \cref{eq:test_loss_single_curve} will be either monotonically decreasing, U-shaped or monotonically increasing in $t$. We can see this by noting that under the aforementioned conditions, the first time derivative of $\lossgeni(t)$ changes signs only at one point, namely at $\mzt = \truez$. If $\truez \geq \starz$ or $\truez \leq \modelz(0)$, this point is never passed, and the error curve $\lossgeni(t)$ is monotonically decreasing or increasing, respectively, in $t$. If $\starz > \truez > \modelz(0)$, the point $\truez$ is passed once, and the error curve $\lossgeni(t)$ is U-shaped. Therefore, epoch-wise double descent in the decoupled multi-layer model, if it appears, is a result of a superposition of monotonically decreasing, U-shaped or monotonically increasing curves.  

Next, we find potential inflection points of the error curve, $\lossgeni(t)$, by solving
\begin{align*}
    \frac{d^2\lossgeni(t)}{dt^2}  = 0.
\end{align*}
Solutions to this equation include $\mzt=0$ and $\mzt=\starz$, none of which are inflection points, as $\mzt \in (0, \starz]$. The other two solutions are
\begin{align*}
    \hat{\modelz}^{\pm} = \frac{ \lambdayx \big( 6 - 3\rho \pm \sqrt{9 \rho^2-4 \rho+4}\big) }{8 \lambda}
\end{align*}
For our further analysis, we exclude the case $\rho=1$, as it under \assump would require $\modelz(0)=0$, contradicting the condition $\modelz(0) > 0$ and making the weight $\mzt$ inactive. For $\rho \in [0, 1)$, we can verify that
\begin{align*}
     1-\rho > \frac{6 - 3\rho - \sqrt{9 \rho^2-4 \rho+4} }{8} > 0,
\end{align*}
and, therefore, $\hat{z}^{-} \in (0, \truez)$. We can also verify, for $\rho \in (0, 1)$, that
\begin{align*}
    1 >\frac{ 6 - 3\rho + \sqrt{9 \rho^2-4 \rho+4} }{8} > 1-\rho,
\end{align*}
and, hence, $\hat{z}^{+} \in (\truez, \starz)$. For $\rho=0$, we have $\hat{z}^{+} = \starz$, and $\hat{z}^{+}$ is therefore not an inflection point in this case.

We evaluate the second derivative at a point in between the two roots, to find
\begin{align*}
     \frac{d^2}{dt^2}\lossgeni(t)\Big|_{\mzt = \frac{(6-3\rho)\lambdayx}{8\lambda}} &= -\frac{27 L^2 \bar{\lr}^2 \lambdayx^6 (\rho-2)^3 (3\rho+2) \big(9 \rho^2-4 \rho+4\big) }{131,072 \lambda^3}  > 0,
\end{align*}
i.e. the error curve $\lossgeni(t)$ is convex in between the roots $\hat{z}^{-}$ and $\hat{z}^{+}$. For $\rho \in [0, 1)$, we additionally evaluate the second derivative at a point in the interval $(0, \hat{z}^{-})$, for which
\begin{align*}
    \frac{d^2}{dt^2}\lossgeni(t)\Big|_{\mzt = \frac{(3-3\rho)\lambdayx}{8\lambda}} = -\frac{27 L^2 \bar{\lr}^2 \lambdayx^6 (\rho - 1)^4 (3\rho + 5) (9 \rho + 5) }{131,072 \lambda^3} < 0.
\end{align*}
Moreover, for $\rho \in (0, 1)$, we evaluate the second derivative at a point in the interval $(\hat{z}^{+}, \starz)$, for which we find
\begin{align*}
       \frac{d^2}{dt^2}\lossgeni(t)\Big|_{\mzt = \frac{(8-\rho)\lambdayx}{8\lambda}} = -\frac{ L^2 \bar{\lr}^2 \lambdayx^6 \rho^2(\rho -8)^3  (5 \rho - 12)}{131,072 \lambda^3} < 0.
\end{align*}
Hence, we find that the error curve changes convexity at the roots $\hat{z}^{-}$ and $\hat{z}^+$, and that these roots are indeed inflection points under the given restrictions on $\rho$. Remember that under \assump with $\bar{\lr}, \, \modelz(0) > 0$, an active weight $\mzt$ following the dynamics in \cref{eq:deeper_models_dynamics_z}, will monotonically increase in $t$ (as concluded above). Hence, the inflection points $\hat{z}^{\pm}$ will correspond to unique time points $\hat{t}^{\pm}$, with $t^-$ an inflection point provided also that $\modelz(0) < \hat{z}^-$. 

To conclude, with $\mzt$ following the dynamics in \cref{eq:deeper_models_dynamics_z}, under \assump and with $\bar{\lr}, \modelz(0) > 0$, the error curve $\lossgeni(t)$ will have up to two inflection points. If $\modelz(0) < \hat{z}^-$ and $\rho \in [0, 1)$, there will be one inflection point, $t^-$, lying on the interval $(0, t^{(1-\rho)\starz})$. Moreover, if $\rho \in (0, 1)$, there will be one inflection point, $t^+$, lying on the interval $(t^{(1-\rho)\starz}, \infty)$.

\end{document}